\documentclass{article}

\usepackage[main, final]{neurips_2026}

\usepackage[utf8]{inputenc} 
\usepackage[T1]{fontenc}    
\usepackage{hyperref}       
\usepackage{url}            
\usepackage{booktabs}       
\usepackage{amsfonts}       
\usepackage{amssymb}
\usepackage{nicefrac}       
\usepackage{microtype}      
\usepackage{xcolor}         
\usepackage{amsmath}
\usepackage{graphicx}
\usepackage{mathtools}
\usepackage{tabularx}
\usepackage{caption}
\usepackage{float}
\usepackage{subcaption}
\usepackage{multirow}
\usepackage{wrapfig}
\title{Quadratic Bezier Splatting with Structural Initialization for Efficient Painting Rendering}

\author{
Jinfan Liu$^{\dagger}$ \quad
Xuhan Zhan \quad
Mengqin Zhao \quad
Guangyi Deng \quad
Wuze Zhang \quad
Bingbing Ni$^{\dagger}\thanks{Corresponding author.}$ \\
{\small Shanghai Jiao Tong University \quad $\dagger$ Guangdong Laboratory of Artificial Intelligence and Digital Economy (SZ)} \\
}

\begin{document}

\maketitle
\begin{figure}[h!]
\centering
\includegraphics[width=1.0\linewidth]{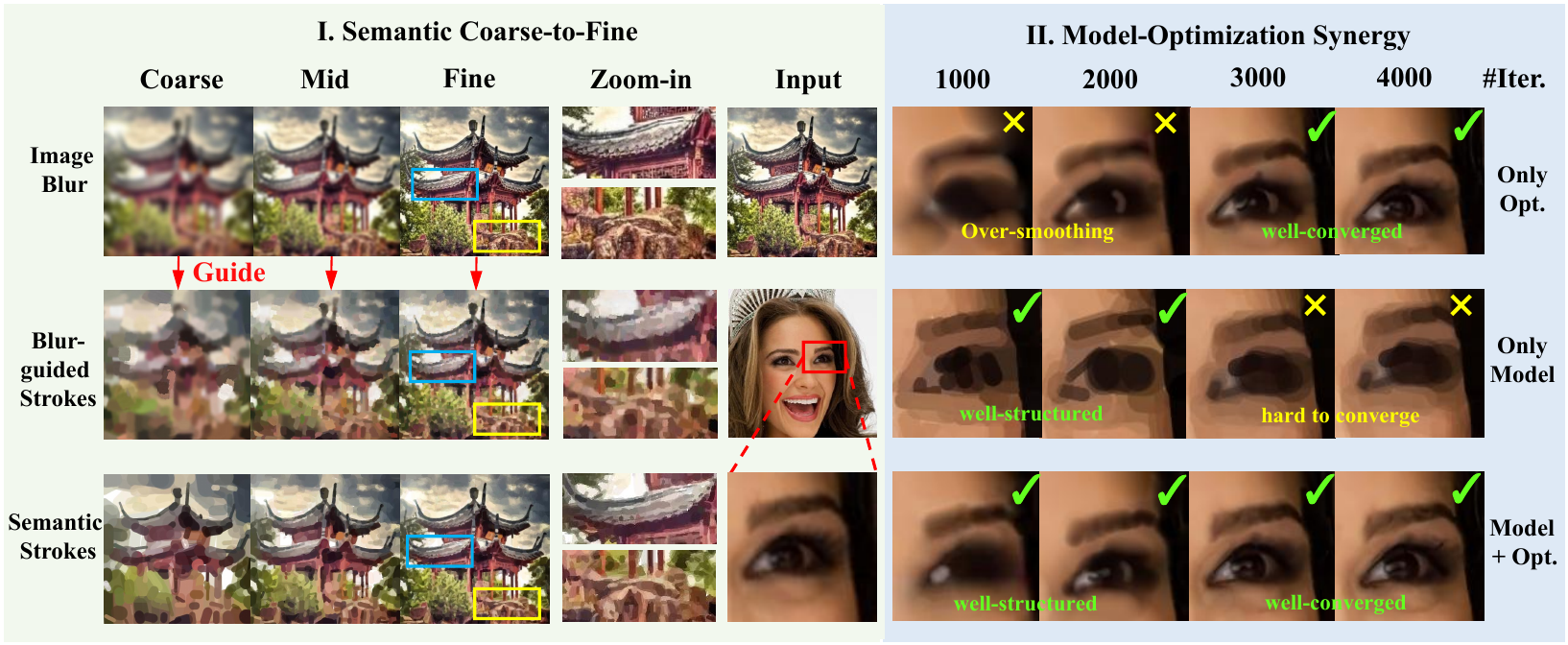}
\caption{
\textbf{Motivation.}
\textit{Left:} Image-blur-based coarse-to-fine optimization mainly performs
low-level smoothing and produces fragmented stroke layouts, while our semantic
initialization yields more coherent multi-scale strokes.
\textit{Right:} Optimization alone can eventually converge but suffers from
early over-smoothing, whereas model prediction alone provides good early
structure but is difficult to refine. Combining semantic initialization with
optimization achieves both well-structured strokes and stable convergence.
}
\label{fig:motivation}
\end{figure}

\begin{abstract}

Stroke-based rendering is often optimized as a pixel-level reconstruction task. This over-reliance on global error reduction not only degrades strokes into fragmented curves due to a lack of structured planning, but also allows locally defective primitives to be concealed by overlapping layers. To address these issues, we introduce a model-guided framework for stroke-audited rendering. Our framework uses Quadratic Bezier Splatting to analytically model continuous curved strokes with efficient differentiable alpha compositing. Built on this renderer, a feedforward multi-scale stroke decoder provides a structural initialization with learned stroke-scale allocation, while test-time stroke optimization supplies the reconstruction capacity needed for high-fidelity refinement. We further propose a single-stroke audit mechanism that explicitly evaluates local stroke quality, preventing defective primitives from surviving optimization through layer-wise compensation. Compared with existing methods, our approach achieves coherent layouts, improves PSNR by up to 5.46 dB, yields a 16\% more compact representation, and accelerates combined renderer forward and backward computation by 220.3$\times$.
\end{abstract}

\section{Introduction}
\vspace{-0.2cm}
AI-driven visual content creation has evolved from pixel-level synthesis to intelligent \textbf{creative assistance} \cite{im2oil, p-trans}. Stroke-based rendering (SBR) reconstructs images using interpretable brushstrokes rather than unstructured grids \cite{haeberli1990paint, hertzmann1998painterly}, allowing for a more natural expression of artistic intent \cite{meier1996painterly, im2oil, snp, p-trans}. However, effective SBR requires more than global pixel matching; it demands structured stroke planning, where each primitive coherently explains a local region with plausible geometry and direction \cite{hertzmann1998painterly, hertzmann2001paint}.

Existing SBR pipelines often lose the structure due to three limitations. First, discontinuous representations approximate curves with discrete segments~\cite{im2oil,snpsd,snp,p-trans,p-rl}, creating inherently fragmented layouts. Second, progressive Gaussian smoothing~\cite{im2oil,snp} yields low-frequency targets rather than structural plans, ignoring boundary continuity. Third, over-reliance on global reconstruction loss allows defective strokes to persist, as local errors are masked by overlapping layers. Consequently, painting degenerates into unstructured pixel accumulation rather than coherent planning.

To address these issues, we propose \textbf{Stroke-Audited Quadratic Bezier Splatting}, a model-guided optimization framework integrating a curve-native differentiable renderer, a feedforward structural initializer, and a local stroke-auditing optimizer. This hybrid design bridges the gap between feedforward prediction (providing a structural lower bound) and differentiable optimization (achieving a high-fidelity upper bound), overcoming single-paradigm limitations. Specifically, our renderer models open brushstrokes as continuous quadratic Bezier curves using an efficient splatting formulation. By computing analytic point-to-curve distances rather than discretizing curves into segments, it achieves a compact, continuous representation efficient for joint optimization. Built upon this renderer, a multi-scale stroke decoder initializes Bezier strokes, replacing heuristic frequency abstraction with a learned scale allocation. This yields a structurally coherent starting layout, significantly reducing the optimization burden of discovering stroke organization from random or purely signal-level initializations. Fig.~\ref{fig:motivation} illustrates our pipeline and its comparison with existing SBR paradigms.

Finally, we make test-time optimization locally accountable through single-stroke auditing. 
Rather than relying solely on global image error, our audit explicitly evaluates the local contribution of each visible stroke to drive adaptive stroke scheduling, preventing flawed primitives from being masked by overlapping strokes. Evaluations show up to 5.46 dB higher PSNR and a 16\% more compact representation than DP. Combined renderer forward and backward time is 220.3$\times$ faster than DiffVG, while a user study yields higher mean ratings for visual coherence and organization than representative SBR baselines.

Our main contributions are threefold: 1) we introduce Quadratic Bezier Splatting, a differentiable renderer that models open brushstrokes as continuous Bezier primitives via analytic point-to-curve distances, avoiding discrete segment approximations while enabling efficient splatting; 2) we propose a feedforward multi-scale stroke decoder as a structural initializer, replacing heuristic frequency-level image blurring with a learned stroke-scale allocation; and 3) we develop a single-stroke audit mechanism that evaluates the local contribution of each primitive (e.g., purity, benefit density, and residual demand) to guide adaptive deletion, splitting, and respawning during optimization.

\vspace{-0.3cm}
\section{Related Works}

\paragraph{Differentiable Curve Rendering.}
Stroke-based rendering represents images as ordered brush marks~\cite{curtis1997watercolor, decarlo2002stylization, haeberli1990paint, hertzmann1998painterly, hertzmann2001paint, kang2009flow, litwinowicz1997processing, meier1996painterly, winnemoller2006realtime}.
Most existing methods approximate strokes as piecewise line segments, causing fragmentation and hindering end‑to‑end differentiability.
Recent differentiable vector graphics~\cite{li2020diffvg} enabled gradient‑based optimization, advancing differentiable SVG, deep vectorization, and text‑guided sketch synthesis~\cite{carlier2020deepsvg, egiazarian2020deep, frans2022clipdraw, jain2023vectorfusion, reddy2021im2vec, xing2023diffsketcher, chen2024vectorization, liu2025bezier}.
Even semantic‑aware methods~\cite{segpainting2024, hu2024artist} still rely on discrete line primitives, preventing continuous curve refinement.
Furthermore, existing differentiable renderers either target filled shapes or sample curves into intermediate polylines~\cite{liu2026dualstroke}, introducing angular points that disrupt per‑stroke analysis and gradient flow.
The closest to ours is Bezier Splatting~\cite{liu2025bezier}, which splats Bezier curves for general vector graphics by sampling them into 2D Gaussians.
Its open mode places Gaussians along a centerline, so overlapping samples do not guarantee uniform coverage across the stroke width and the contour can fluctuate; its closed mode fills a region bounded by two curves, which tends to form blobs or solid patches rather than single-skeleton brush trajectories.
Such sampling also multiplies the number of runtime primitives per stroke by two to three orders of magnitude (Sec.~\ref{sec:renderer_analysis}).
To overcome these limitations, we directly render open quadratic Bézier strokes through an analytical point‑to‑curve distance, achieving fully differentiable, fragmentation‑free continuous curves inspired by the splatting paradigm~\cite{kerbl20233dgs}.

\paragraph{Structural Initialization.}
Traditional pipelines simulate coarse-to-fine painting by applying progressively decreasing Gaussian blur to the target image~\cite{im2oil, snp}, but this isotropic smoothing is a fixed signal-level abstraction that cannot distinguish object layout, boundary continuity, or paintable regions.
Besides, although neural stroke predictors which include reinforcement-learning agents~\cite{p-rl}, feed-forward set predictors~\cite{snpsd, p-trans, tang2025attentionpainter}, and program-induction models~\cite{ganin2018spiral, ha2018sketchrnn}, can produce globally plausible layouts, their outputs are often raster images or heterogeneous stroke representations that cannot directly initialize the geometric parameters of a Bezier splatting optimizer.
To address this limitation, we introduce a learned predictor with multi-scale stroke queries that provides a structurally organized warm start for the differentiable optimizer, supplying structure-aware geometric initialization unavailable from blur-based schedules.

\paragraph{Stroke Quality Evaluation.}
Rendering quality is conventionally assessed through composite metrics such as SSIM and LPIPS~\cite{wang2004ssim, zhang2018lpips}, supplemented by low-level edge, contour, and saliency cues~\cite{canny1986edge, itti1998saliency, kass1988snakes}.
However, these measures capture only overall image similarity or pixel-level importance, offering no insight into the contribution of individual strokes.
A comprehensive survey of stylization~\cite{kyprianidis2013state} and recent neural stroke rendering methods~\cite{nolte2023survey} significantly advance visual fidelity, yet both paradigms evaluate only the final composite canvas without diagnosing per-stroke quality.
Consequently, a locally destructive stroke can hide beneath subsequent layers, satisfying global metrics while degrading the underlying stroke layout.
We address this gap by introducing a per-stroke quality evaluation framework that explicitly assesses individual contributions, enabling the optimizer to identify harmful strokes and reallocate capacity to structurally significant regions.

\begin{figure*}[t]
    \centering
    \includegraphics[width=0.98\textwidth]{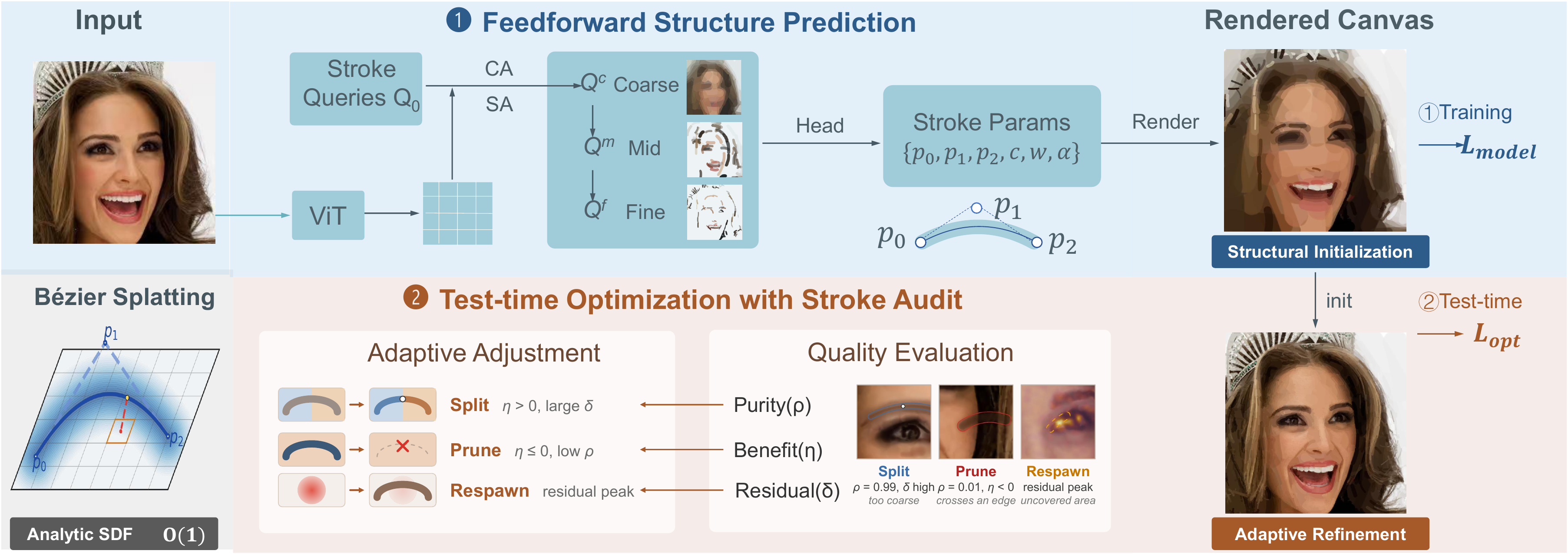}
    \caption{
\textbf{Framework.}
The Input image is represented by quadratic B\'{e}zier strokes rendered through an analytic SDF-based differentiable splatting renderer.
A ViT-based feedforward predictor decodes multi-scale stroke queries into coarse, mid, and fine stroke groups, yielding stroke parameters as a structural initialization trained with $\mathcal{L}_{\text{model}}$.
Starting from this initialization, test-time optimization refines the canvas with $\mathcal{L}_{\text{opt}}$.
In parallel, the stroke audit module measures each stroke's purity, benefit density, and residual density, and triggers adaptive pruning, splitting, and respawning to reallocate stroke capacity toward unresolved structures.
}
    \label{fig:framework}
\end{figure*}

\vspace{-0.3cm}
\section{Methodology}
Our method is motivated by a key observation: a feedforward stroke model can provide a reliable structural starting point, while test-time differentiable stroke optimization provides the reconstruction capacity needed for high-fidelity painting. 
Given a target image $I_{gt}$, our goal is to find a set of strokes $\mathcal{S}=\{S_i\}$ that minimizes reconstruction error while keeping individual strokes locally constructive and structurally coherent.

As illustrated in Figure \ref{fig:framework}, our core framework consists of three components: 1) a differentiable quadratic Bezier splatting renderer that supports free-form strokes; 2) a feedforward multi-scale stroke decoder that generates a structural initialization rather than the final painting; and 3) adaptive test-time optimization guided by single-stroke audit metrics, making local stroke quality explicit during refinement.

\subsection{Quadratic Bezier Splatting}
We adopt quadratic Bezier curves as stroke primitives. Each stroke is parameterized as $S_i = (P_0^i, P_1^i, P_2^i, \mathbf{c}_i, w_i, \alpha_i)$, where $P_0^i, P_1^i, P_2^i \in \mathbb{R}^2$ are the start, control, and end points, $\mathbf{c}_i \in [0,1]^3$ the RGB color, $w_i \in \mathbb{R}^+$ the half-width, and $\alpha_i \in [0,1]$ the opacity.
The curve is defined as:
\begin{equation}
\mathbf{B}_i(t) = (1 - t)^2 P_0^i + 2(1 - t)t P_1^i + t^2 P_2^i, \quad t \in [0,1].
\end{equation}

This formulation provides a compact, differentiable representation with continuous tangent direction along the stroke, enabling stable optimization of smooth, coherent brush trajectories.

Inspired by the splatting mechanism in 3D Gaussian Splatting, we render strokes via 
per-pixel $\alpha$-compositing, which is GPU-friendly for high-speed parallel 
rasterization. A sigmoid kernel $\sigma$ is adopted to map distance to opacity, 
producing smooth, anti-aliased stroke boundaries with continuous gradients for 
end-to-end optimization. For a queried pixel $\mathbf{p}$, we consider the set of 
overlapping strokes $\mathcal{N}(\mathbf{p})$ within their local support, sort them in 
front-to-back order, and accumulate their contributions as:
\begin{equation}
I(\mathbf{p}) = \sum_{i \in \mathcal{N}(\mathbf{p})} 
\left( \prod_{j<i} \bigl(1 - a_j(\mathbf{p})\bigr) \right) 
\mathbf{c}_i\, a_i(\mathbf{p}), \quad
a_i(\mathbf{p}) = \alpha_i \cdot 
\max\!\left(0,\;
\frac{\sigma\!\left(\frac{w_i - D_i(\mathbf{p})}{\tau w_i}\right) - 0.5}
     {\sigma(1/\tau) - 0.5}
\right),
\end{equation}
where $\tau$ controls edge softness, and $D_i(\mathbf{p})$ is the shortest distance 
from $\mathbf{p}$ to the quadratic Bezier curve, computed exactly and analytically 
in constant time via solving a univariate cubic equation; see supplementary for the full derivation~\ref{sec:analytic_distance_quadratic_bezier}.

\subsection{Feedforward Structural Initialization}
To provide a structural initialization for subsequent refinement, we employ a feedforward transformer that predicts a dense set of parametric strokes. To efficiently handle high-resolution inputs, this predictor operates in a patch-wise manner. The input image is divided into independent local patches, and each patch is encoded by a Vision Transformer backbone into spatial tokens \(F\), discarding the class token to preserve spatial locality.

\textbf{Multi-Scale Stroke Queries.}
We partition \(N\) learnable stroke queries per local patch into coarse, mid, and fine groups, denoted \(\mathcal{Q} = (\mathcal{Q}^c, \mathcal{Q}^m, \mathcal{Q}^f)\) with dimensions \(\mathbb{R}^{N_c \times d}, \mathbb{R}^{N_m \times d}, \mathbb{R}^{N_f \times d}\) and ratio \(N_c:N_m:N_f = 2:3:5\). Each group is assigned a scaled geometric prior with bounded sizes via its initial stroke template \(S_i^0\), targeting large color regions, boundary transitions, and local residual hints respectively; see template details in Suppl.~\ref{tab:template_priors}. With learnable group-specific embeddings \(E^{\{c,m,f\}} \in \mathbb{R}^{N_{\{c,m,f\}} \times d}\) and a shared sinusoidal position encoding \(\mathrm{PE}\) computed via the local stroke index, the initial queries are:
\begin{equation}
\mathcal{Q}_0 = (E^c + \mathrm{PE}^c,\; E^m + \mathrm{PE}^m,\; E^f + \mathrm{PE}^f)
\end{equation}

\textbf{Multi-Scale Stroke Decoding.}
Queries are decoded in a cascaded manner, each stage aggregating image features via cross-attention \(\operatorname{CA}(Q, KV)\) followed by self-attention:
\begin{equation}
(\tilde{\mathcal{Q}}^c, \tilde{\mathcal{Q}}^m, \tilde{\mathcal{Q}}^f) = \big(
\operatorname{SA}(\operatorname{CA}(\mathcal{Q}_0^c, F)),\;
\operatorname{SA}(\operatorname{CA}(\mathcal{Q}_0^m, [F; \tilde{\mathcal{Q}}^c])),\;
\operatorname{SA}(\operatorname{CA}(\mathcal{Q}_0^f, [F; \tilde{\mathcal{Q}}^c; \tilde{\mathcal{Q}}^m]))
\big)
\end{equation}
Each decoded query \(\tilde{q}_i \in \tilde{\mathcal{Q}} = [\tilde{\mathcal{Q}}^c; \tilde{\mathcal{Q}}^m; \tilde{\mathcal{Q}}^f]\) produces bounded residuals for geometry and color parameters via \(\operatorname{Head}(\cdot) = \gamma \tanh(\operatorname{MLP}(\cdot))\), which refine the initial template \(S_i^0\) as follows:
\begin{equation}
S_i = \operatorname{clamp}\!\big(S_i^0 + \operatorname{Head}(\tilde{q}_i),\; 0,\; 1\big)
\end{equation}

The network is trained end-to-end through the differentiable Bezier
splatting renderer, with \(\mathcal{L}_{\text{model}} = \|\hat{I}-I_{gt}\|_1 + \lambda_{\ell_2} \|\hat{I}-I_{gt}\|_2^2 + \lambda_{\text{edge}} \|\Delta \hat{I} - \Delta I_{gt}\|_1\), which combines pixel-wise reconstruction with a Laplacian edge loss to preserve salient contours.

\subsection{Stroke Auditing and Adaptive Optimization}
\label{sec:stroke_auditing}
While the feedforward model provides a structurally meaningful initialization, optimizing solely with a global reconstruction loss fails to evaluate the individual contribution of each stroke. A locally destructive stroke can be concealed by subsequent overlapping layers, escaping detection while compromising local painting quality. To address this, we introduce single-stroke auditing metrics that explicitly quantify per-stroke quality, coupled with an adaptive scheduling strategy that dynamically refines the stroke population.

For stroke $S_i$, its effective visible contribution at pixel $\mathbf{p}$ under front-to-back compositing is defined as follows, where $\widetilde{A}_i$ denotes the weighted supporting area of stroke $S_i$:
\begin{equation}
\tilde{\phi}_i(\mathbf{p}) = a_i(\mathbf{p}) \prod_{j<i}\bigl(1-a_j(\mathbf{p})\bigr),\quad
\tilde{A}_i = \sum_{\mathbf{p}} \tilde{\phi}_i(\mathbf{p}).
\end{equation}

Let $\hat{I}^{(i-1)}$ and $\hat{I}^{(i)}$ denote the canvas before and after compositing $S_i$, and $\hat{I}$ denote the final canvas. We compute three quantities online during rasterization, which are accumulated directly in the renderer to avoid explicit leave-one-out re-rendering:
\begin{equation}
\Delta_i^+ = \sum_{\mathbf{p}} \max\!\Big(0,\; \|\hat{I}^{(i-1)}(\mathbf{p}) - I_{gt}(\mathbf{p})\|_2^2 - \|\hat{I}^{(i)}(\mathbf{p}) - I_{gt}(\mathbf{p})\|_2^2\Big),
\end{equation}
\begin{equation}
\Delta_i^- = \sum_{\mathbf{p}} \max\!\Big(0,\; \|\hat{I}^{(i)}(\mathbf{p}) - I_{gt}(\mathbf{p})\|_2^2 - \|\hat{I}^{(i-1)}(\mathbf{p}) - I_{gt}(\mathbf{p})\|_2^2\Big),
\end{equation}
\begin{equation}
R_i^{\text{supp}} = \sum_{\mathbf{p}} \tilde{\phi}_i(\mathbf{p}) \, \|I_{gt}(\mathbf{p}) - \hat{I}(\mathbf{p})\|_1.
\end{equation}

From these, we derive three auditing metrics to measure stroke purity ($\rho_i$), net benefit per unit effective support ($\eta_i$), and unmet reconstruction demand ($\delta_i$), respectively:
\begin{equation}
\rho_i = \frac{\Delta_i^+}{\Delta_i^+ + \Delta_i^- + \epsilon},\quad
\eta_i = \frac{\Delta_i^+ - \lambda_{\text{bad}} \Delta_i^-}{\tilde{A}_i + \epsilon},\quad
\delta_i = \frac{R_i^{\text{supp}}}{\tilde{A}_i + \epsilon}.
\end{equation}

Every $T$ iterations after a warm-up stage, the system performs adaptive scheduling consisting of three operations. 1) \textbf{Pruning} removes harmful or negligible strokes: those with $\eta_i \leq 0$ and either low purity $\rho_i$ or insufficient effective support $\tilde{A}_i$, as well as strokes with near-zero opacity or excessive radius. Newly edited strokes are temporarily protected from removal. 2) \textbf{Splitting} refines strokes that are beneficial but too coarse: those with $\eta_i > 0$, sufficient $\tilde{A}_i$, high yet non-saturated $\rho_i$, and valid geometry, prioritized by residual density $\delta_i$, and subdivides them at the parametric midpoint via de Casteljau splitting. 3) \textbf{Respawning} reallocates the budget freed by deletion to local maxima of the residual map via non-maximum suppression; new strokes sample target colors and locally aligned directions where reliable, otherwise random orientations. Optimization resumes after each scheduling event, transferring capacity from low-value to structurally significant regions.

Optimization proceeds by minimizing:
\begin{equation}
\mathcal{L}_{\text{opt}} = \|\hat{I} - I_{gt}\|_2^2 + \lambda_{\text{dir}} \mathcal{L}_{\text{dir}},
\end{equation}
where $\mathcal{L}_{\text{dir}}$ encourages alignment between stroke tangents and the image direction field, activated after an early warm-up and applied periodically thereafter; see supplementary for the exact formulation of $\mathcal{L}_{\text{dir}}$~\ref{sec:direction_alignment_loss}. The auditing and scheduling module is integrated as a plug-and-play component orthogonal to the gradient backpropagation flow.

\begin{figure}[h!]
    \centering
    \includegraphics[width=0.99\linewidth]{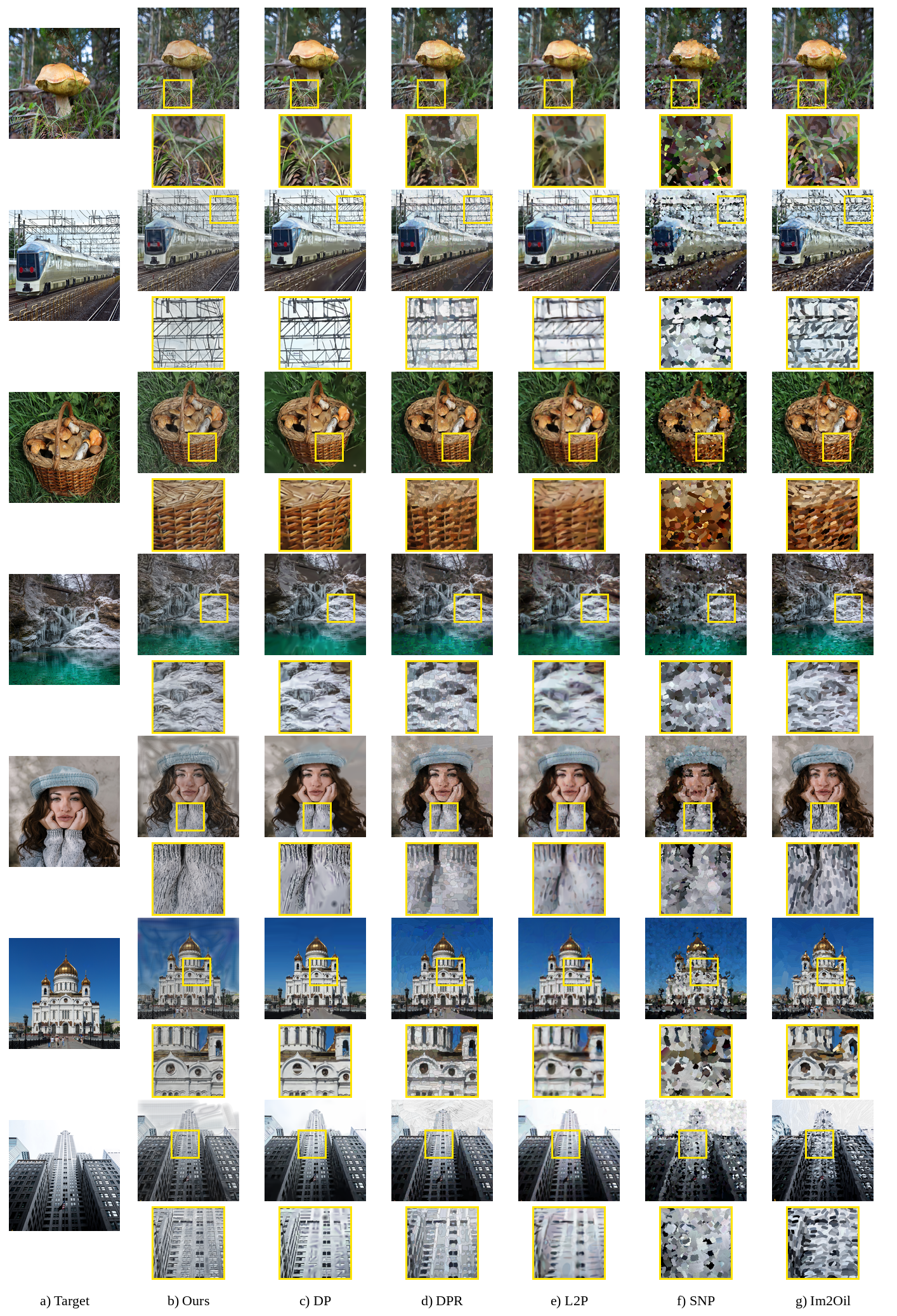}
    \caption{
    \textbf{Qualitative comparison with state-of-the-art methods.}
    Compared with existing approaches, our method produces more coherent structural layouts and better organized stroke flows while preserving cleaner local details.
    Baseline methods often exhibit fragmented strokes, unstable boundaries, and texture-driven artifacts, especially under limited stroke budgets.
    }
    \label{fig:qualitative}
\end{figure}
\vspace{-0.1cm}
\section{Experiments}
\label{sec:exp}

\subsection{Experimental Setup}

\paragraph{Datasets and Metrics.}
We evaluate on DIV2K validation images~\cite{div2k} for natural-image reconstruction and the Im2Oil Gallery dataset~\cite{im2oil} for artistic oil-painting rendering.
The main comparison uses DIV2K at $600 \times 600$ with 5K strokes and Gallery at $600 \times 800$ with 6K strokes.
We report PSNR, SSIM, LPIPS, average time per stroke, and final stroke representation size.
To verify that strokes act as meaningful painting units rather than merely matching pixels globally, our ablations report three stroke-audit metrics (Sec.~\ref{sec:stroke_auditing}): purity $\rho_i$, benefit density $\eta_i$, and residual density $\delta_i$.

\paragraph{Baselines and Implementation.}
We compare with representative SBR methods spanning search-based, feedforward, optimization-based, and hybrid paradigms: Im2Oil~\cite{im2oil}, DPR~\cite{snpsd}, PT~\cite{p-trans}, L2P~\cite{p-rl}, SNP~\cite{snp}, and DP~\cite{liu2026dualstroke}.
When exact stroke budgets are unavailable, we use the closest comparable setting at each resolution.
Our feedforward initializer uses a ViT-small patch encoder, multi-scale stroke queries, a cascaded transformer decoder, and MLP heads, totaling about 28M parameters.
It is trained on Flickr2K for 200 epochs using AdamW with learning rate $1\times10^{-4}$, weight decay $1\times10^{-4}$, and batch size 32 on 4 NVIDIA RTX 4090 GPUs.
All reported inference and renderer timings are measured on a single NVIDIA RTX 4090.
At inference, the initializer processes non-overlapping $256\times256$ patches with at most 250 strokes per patch; test-time optimization adds strokes toward the target budget during 5000 Adam refinement steps with initial learning rate 0.02, cosine decay. We use the rasterization schedule $\tau: 1.0 \rightarrow 0.6 \rightarrow 0.3$ at steps 100 and 200, and $\lambda_{\text{dir}}=0.04$ with 200-step warm-up and 5-step interval. The adaptive auditing module starts at
  step 100, runs every 50 steps, and is disabled in the last 500 steps.
Architecture and training details are provided in Suppl.~\ref{sec:detailed_model_architecture}.

\subsection{Main Results}

Table~\ref{tab:sota_quantitative} compares our method with existing SBR baselines under compact stroke budgets.
Across both DIV2K and Gallery, our method achieves superior reconstruction quality and perceptual fidelity while maintaining a compact final stroke representation.
The gains are especially clear under limited stroke budgets, where existing methods either lose structural detail due to feedforward simplification or spend optimization capacity compensating for fragmented stroke layouts.

Quantitatively, our method achieves 28.52 PSNR, 0.8720 SSIM, and 0.1713 LPIPS on DIV2K, outperforming the strongest available baseline DP by 4.29 dB in PSNR and 0.1873 in SSIM while reducing LPIPS by 0.1267.
On Gallery, the advantage is even larger: our method reaches 32.65 PSNR, 0.8936 SSIM, and 0.1547 LPIPS, improving over DP by 5.46 dB in PSNR and 0.1324 in SSIM, with a 0.0756 LPIPS reduction.
These substantial gains show that our rendered canvases preserve image content, global structure, and perceptual quality more faithfully across both natural and artistic images.

The efficiency results further highlight the fidelity-efficiency balance.
Our method takes 26.9ms per stroke, faster than Im2Oil (32.9ms), DP (64.5ms) and SNP (276.1ms), while achieving much higher reconstruction quality.
The final representation size is 21.2K, close to the lightweight PT representation (18.0K) and smaller than DP (25.2K), L2P (31.4K), DPR (56.4K), Im2Oil (61.0K), and SNP (96.2K).
Our hybrid formulation offers a stronger fidelity-efficiency trade-off: the feedforward initializer provides an organized structural scaffold, and test-time optimization supplies the capacity needed for high-quality refinement.

Qualitative comparisons in Fig.~\ref{fig:qualitative} further support this conclusion.
Our method produces cleaner object boundaries, more coherent long-range stroke flows, and better organized local details.
In contrast, baselines often show fragmented strokes, unstable contours, or texture-driven artifacts, especially in regions where compact stroke budgets require each primitive to explain a meaningful local structure.
These results suggest that continuous Bezier primitives, structural initialization, and stroke-aware optimization are complementary for high-fidelity painting rendering.

\begin{table*}[t]
\centering
\scriptsize
\renewcommand{\arraystretch}{0.95} 
\setlength{\tabcolsep}{3.2pt}
\caption{
Quantitative comparison on DIV2K and Gallery with 5K and 6K strokes, respectively.
Time is average time per stroke; Size is the final stroke representation size.
}
\label{tab:sota_quantitative}

\resizebox{\linewidth}{!}{
\begin{tabular}{c|l|ccc|ccc|cc}
\toprule

\multicolumn{2}{c|}{Dataset}
& \multicolumn{3}{c|}{DIV2K $600\times600$ / 5K}
& \multicolumn{3}{c|}{Gallery $600\times800$ / 6K}
& \multicolumn{2}{c}{Efficiency}
\\

Paradigm
& Method
& PSNR$\uparrow$
& SSIM$\uparrow$
& LPIPS$\downarrow$
& PSNR$\uparrow$
& SSIM$\uparrow$
& LPIPS$\downarrow$
& Time$\downarrow$
& Size$\downarrow$
\\

\midrule

Search
& Im2Oil~\cite{im2oil}
& 16.83 & 0.3602 & 0.4298
& 20.59 & 0.5056 & 0.4623
& 32.9ms & 61.0K
\\

Opt.
& DPR~\cite{snpsd}
& 20.26 & 0.4614 & 0.4363
& 22.48 & 0.5501 & 0.4838
& \underline{6.3ms} & 56.4K
\\

FF
& PT~\cite{p-trans}
& 14.40 & 0.2518 & 0.5084
& 17.22 & 0.4479 & 0.5325
& \textbf{3.1ms} & \textbf{18.0K}
\\

FF
& L2P~\cite{p-rl}
& 20.72 & 0.5020 & 0.5023
& 24.86 & 0.6573 & 0.4068
& 16.8ms & 31.4K
\\

Opt.
& SNP~\cite{snp}
& 14.79 & 0.3787 & 0.5601
& 17.23 & 0.4345 & 0.5936
& 276.1ms & 96.2K
\\

Search+Opt.
& DP~\cite{liu2026dualstroke}
& \underline{24.23} & \underline{0.6847} & \underline{0.2980}
& \underline{27.19} & \underline{0.7612} & \underline{0.2303}
& 64.5ms & 25.2K
\\

\midrule

FF+Opt.
& Ours
& \textbf{28.52}
& \textbf{0.8720}
& \textbf{0.1713}
& \textbf{32.65}
& \textbf{0.8936}
& \textbf{0.1547}
& 26.9ms
& \underline{21.2K}
\\

\bottomrule
\end{tabular}
}

\vspace{-0.6em}
\end{table*}

\subsection{Renderer Analysis}
\label{sec:renderer_analysis}
Table~\ref{tab:efficiency_comparison} evaluates the rendering backends under a 2K-stroke budget.
Quadratic Bezier Splatting takes 1.37ms and 3.59ms for the forward and backward passes, respectively, yielding a 220.3$\times$ speedup over DiffVG (190.6ms forward and 902.3ms backward).
We compare with Bezier Splatting~\cite{liu2025bezier}, the most closely related splatting renderer, after applying its renderer-level optimizations to our implementation.
Because~\cite{liu2025bezier} discretizes each curve into Gaussian samples, a single stroke expands to 192 Gaussians in its open mode (three degree-4 curves) and 2,688 Gaussians in its closed mode (42 curves), whereas our renderer keeps one analytic primitive per stroke.
Using only 0.5\% and 0.04\% of their runtime primitives, our renderer is 1.99$\times$ and 5.96$\times$ faster, and it gains 1.23 dB PSNR over sampling our own strokes into 64 Gaussians each.
Bezier Splatting reaches higher PSNR by spending orders of magnitude more primitives, reflecting a fidelity-efficiency trade-off; moreover, its sampled or region-filled curves do not preserve the single-skeleton structure of an open brushstroke, which SBR requires.

\begin{figure*}[t]
    \centering
    \includegraphics[width=\linewidth]{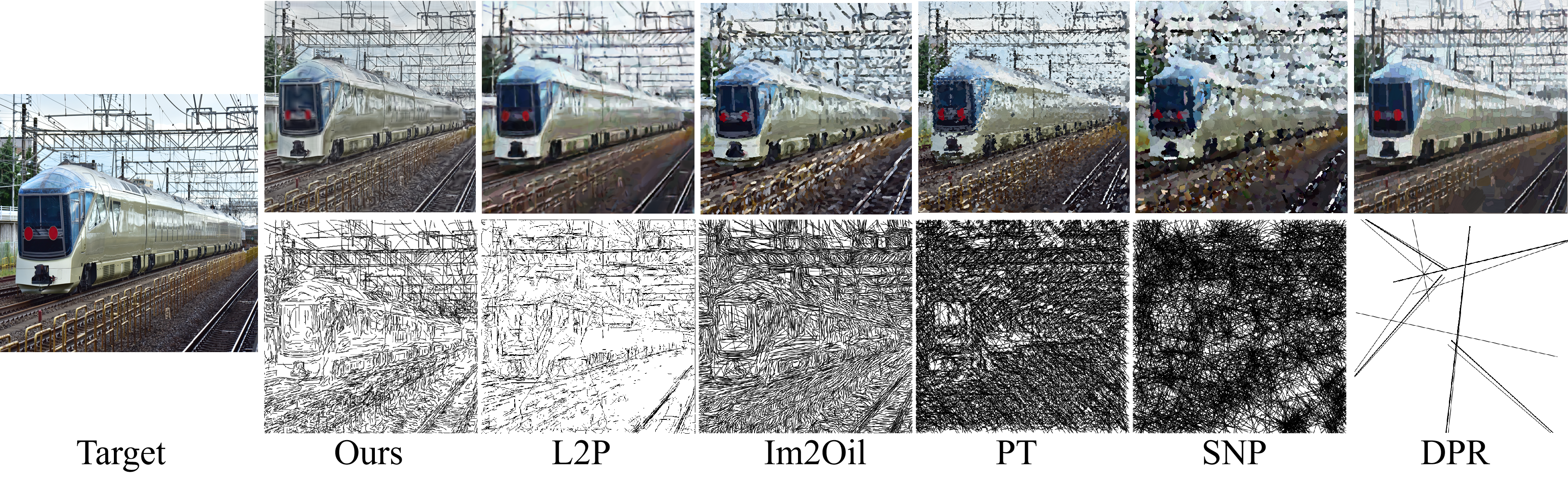}
    \caption{
    \textbf{Curve quality comparison.}
    Polyline-based representations approximate curved strokes using piecewise-linear segments, while our analytic quadratic Bezier splatting directly represents continuous stroke trajectories, yielding strokes with coordinated structural semantics.
    }
    \label{fig:bezier_curve_quality}
\vspace{-0.2cm}
\end{figure*}

Fig.~\ref{fig:bezier_curve_quality} illustrates why this matters geometrically: polyline strokes approximate curvature with segments, whereas quadratic Bezier splatting directly optimizes a continuous trajectory to achieve structurally coordinated strokes.

\begin{table}[t]
\centering
\caption{
Comparison with differentiable rendering methods under a 2K-stroke budget.
\#Prim. is the number of runtime primitives; speedup is relative to the total forward and backward time of DiffVG.
Our implementation applies the renderer-level optimizations of Bezier Splatting~\cite{liu2025bezier} (forward-state reuse, warp-level gradient reduction, and analytical tile culling); Bezier Splatting rows use its released configuration.
Best and second best timing, speedup, and PSNR values are marked among the listed methods.
}
\resizebox{0.95\linewidth}{!}{
\begin{tabular}{lccccc}
\toprule

Method
& \#Prim.
& Fwd(ms)
& Bwd(ms)
& Speedup
& PSNR
\\

\midrule

DiffVG for Quadratic Bezier & 2K & 190.6 & 902.3 & 1.0$\times$ & 27.91 \\
Neural Render for Line & 2K & 312.4 & 1488.7 & 0.6$\times$ & 25.84 \\
Polyline Splatting & 2K & 4.1 & 16.8 & 52.2$\times$ & 29.74 \\

Quadratic Bezier, sampled (64 Gaussians/stroke)
& 128K
& \underline{2.34}
& \underline{3.90}
& \underline{175.1$\times$}
& 28.96
\\

Bezier Splatting, open~\cite{liu2025bezier}
& 384K
& 4.81
& 5.05
& 110.8$\times$
& \textbf{35.87}
\\

Bezier Splatting, closed~\cite{liu2025bezier}
& 5.38M
& 15.76
& 13.81
& 37.0$\times$
& \underline{32.14}
\\

Quadratic Bezier Splatting
& 2K
& \textbf{1.37}
& \textbf{3.59}
& \textbf{220.3$\times$}
& 30.19
\\

\bottomrule
\end{tabular}
}

\label{tab:efficiency_comparison}
\end{table}

\subsection{Ablation Study}

Table~\ref{tab:ablation} studies the main design choices on DIV2K: the feedforward structural initializer, the multi-scale query design, and the adaptive stroke-auditing loop.
The ablations are intended to separate image-level reconstruction quality from stroke-level usefulness, since a globally accurate canvas can still contain redundant or locally defective primitives.

\begin{figure*}[t]
    \centering
    \includegraphics[width=0.9\linewidth]{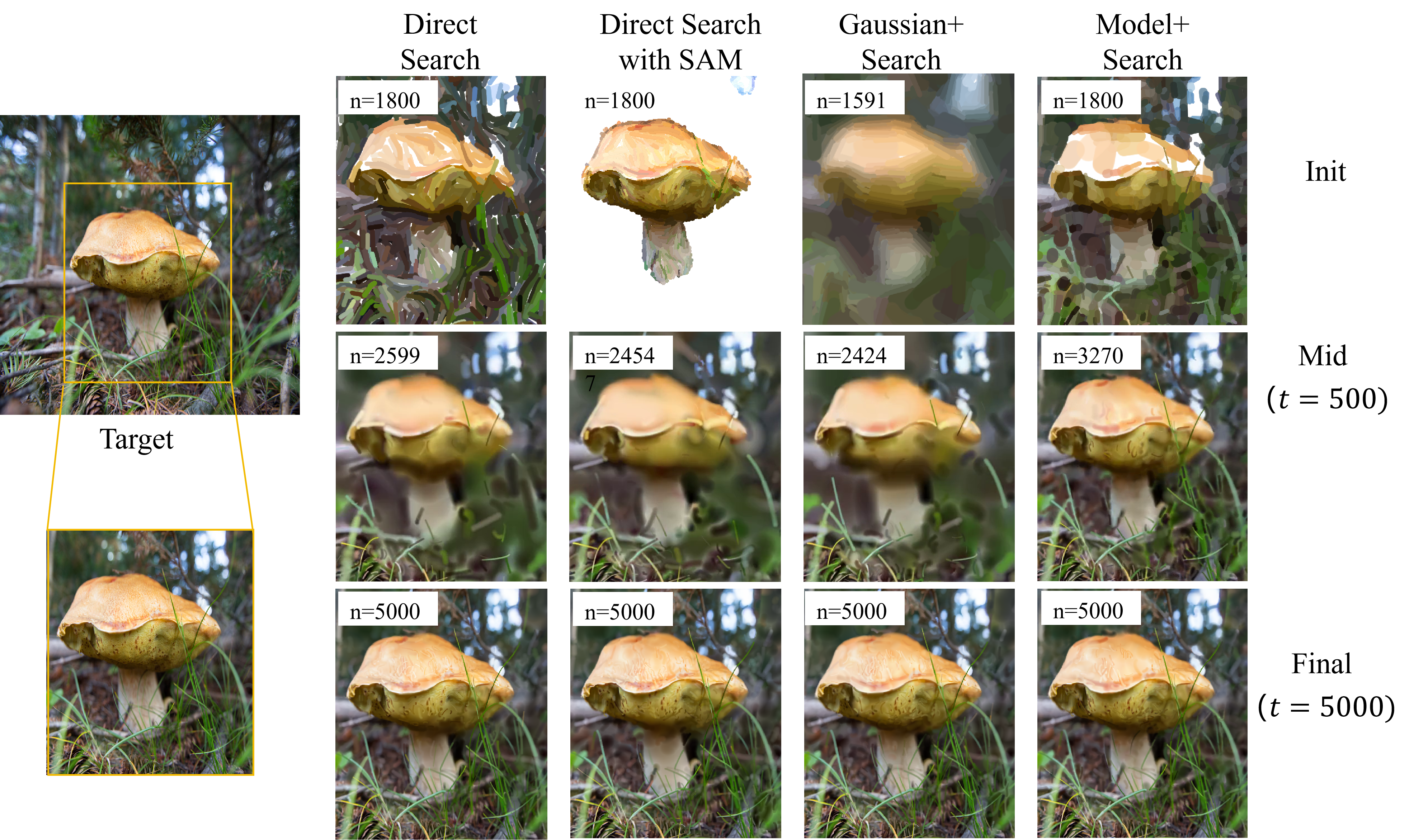}
    \caption{
    \textbf{Initialization comparison.}
    Gaussian coarse-to-fine initialization mainly provides low-frequency smoothing and often produces fragmented stroke layouts after optimization.
    In contrast, our feedforward initialization generates more coherent structural organization, leading to fewer dead strokes, faster convergence, and better final reconstruction quality.
    }
    \label{fig:initialization_comparison}
\end{figure*}

\paragraph{Structural Initialization.}
The raw feedforward output (\textbf{w/o Opt}) provides a structured but low-fidelity canvas, showing that test-time refinement is still needed.
For the additional matched-budget FF-only evaluation, 5,000 feedforward-initialized strokes are directly rendered without continuous optimization or adaptive scheduling. This reaches only 20.58 dB PSNR, supporting the role of the model as a structural warm start.
It is also an efficient one: the initializer runs in 0.15s, whereas gradient-based optimization from random initialization needs over 600 iterations ($\sim$3s) to match the PSNR of the native feedforward output and still yields over-smoothed structure in early iterations (Fig.~\ref{fig:motivation}).
Optimization-only variants such as \textbf{Random Init} and \textbf{Edge Init} can reduce pixel error, but they require the optimizer to discover stroke organization from weaker priors.
Since SBR is ill-posed, with substantially different stroke organizations producing similar raster reconstructions, close image-level scores do not imply equally organized strokes; the stroke-level metrics expose this difference.
Compared with \textbf{Random Init}, the full model improves PSNR by 0.32 dB, SSIM by 0.0225, and reduces residual density from 0.0732 to 0.0630.
Compared with \textbf{Edge Init}, it improves PSNR by 0.17 dB, SSIM by 0.0240, and reduces residual density from 0.0743 to 0.0630.
As shown in Fig.~\ref{fig:initialization_comparison}, the learned structural prior gives a more coherent starting layout, reducing the burden of spatial discovery during subsequent optimization.

\begin{table*}[t]
\renewcommand{\arraystretch}{0.95} 
\setlength{\tabcolsep}{3.2pt}
\centering
\caption{
Ablation study on structural initialization, multi-scale query design, and adaptive stroke auditing. 
``w/o Opt'' denotes the raw output of the feedforward initializer before test-time optimization, using either its native stroke count or the same 5K-stroke budget as the full model. Missing or non-applicable metrics across different testing regimes are denoted by ``/''.
}
\label{tab:ablation}
\resizebox{1\linewidth}{!}{
\begin{tabular}{l|ccc|ccc|cc}
\toprule
Method
& PSNR$\uparrow$
& SSIM$\uparrow$
& LPIPS$\downarrow$
& Purity$\uparrow$
& Benefit$\uparrow$
& Residual$\downarrow$
& Time$\downarrow$
& Size$\downarrow$
\\
\midrule
Full (FF + Opt)
& \textbf{28.52}
& \textbf{0.872}
& \textbf{0.171}
& \textbf{0.9949}
& 1.5496
& \textbf{0.0630}
& 26.9ms
& 21.2K
\\
\midrule
\multicolumn{9}{l}{\textit{Initialization \& Architecture}} \\
\midrule
w/o Opt (FF Only)
& 15.65 & 0.3098 & 0.6028 & 0.7042 & 0.4488 & 0.3502 & / & 1.0K \\
w/o Opt (FF Only, 5K strokes)
& 20.58 & 0.5313 & 0.4067 & 0.9209 & 1.1297 & 0.0739 & / & 21.2K \\
Random Init (Opt Only)
& 28.20 & 0.8495 & 0.1867 & 0.9912 & 1.5381 & 0.0732 & 26.6ms & 21.2K \\
Edge Init (Heuristic)
& 28.35 & 0.8480 & 0.1875 & 0.9924 & 1.5403 & 0.0743 & 26.9ms & 21.2K \\
\midrule
\multicolumn{9}{l}{\textit{Multi-scale Query Design}} \\
\midrule
Single-scale
& 27.03 & 0.826 & 0.203 & 0.9824 & 1.5034 & 0.0792 & 26.6ms & 20.7K \\
Ratio 1:1:1
& 27.14 & 0.831 & 0.196 & 0.9902 & 1.5232 & 0.0786 & 26.7ms & 20.9K \\
w/o CrossAttn
& 27.18 & 0.835 & 0.199 & 0.9801 & 1.4868 & 0.0802 & 26.5ms & 21.0K \\
\midrule
\multicolumn{9}{l}{\textit{Adaptive Stroke Auditing}} \\
\midrule
w/o Audit
& 26.80 & 0.807 & 0.207 & 0.9907 & \textbf{1.7390} & 0.0787 & 26.9ms & 21.1K \\
Prune Only
& 25.93 & 0.772 & 0.256 & 0.9862 & 1.6017 & 0.0773 & 26.9ms & 21.1K \\
Split Only
& 27.33 & 0.831 & 0.178 & 0.9899 & 1.7362 & 0.0762 & 26.9ms & 21.1K \\
Respawn Only
& 27.71 & 0.833 & 0.174 & 0.9907 & 1.6157 & 0.0748 & 26.9ms & 21.1K \\
\bottomrule
\end{tabular}
}
\vspace{-1em}
\end{table*}

\paragraph{Multi-scale Stroke Queries.}
The multi-scale query design allocates stroke capacity across coarse regions, medium-scale boundaries, and fine residual details.
Removing this design with \textbf{Single-scale}, changing the query ratio to \textbf{Ratio 1:1:1}, or removing cross-attention consistently weakens structural and stroke-level metrics in Table~\ref{tab:ablation}.
The full model improves over these variants by 1.49, 1.38, and 1.34 dB in PSNR, respectively, and also yields the best purity (0.9949) and lowest residual density (0.0630).
This supports the use of scale-aware query groups for organized stroke allocation.

\paragraph{Stroke Auditing.}
The audit loop evaluates each visible stroke through purity, benefit density, and residual demand, then reallocates capacity by pruning, splitting, and respawning.
Without auditing, optimization relies more heavily on global image loss and leaves more residual demand on visible stroke support.
The full audit loop improves over \textbf{w/o Audit} by 1.72 dB PSNR and 0.065 SSIM, reduces LPIPS from 0.207 to 0.171, and lowers residual density from 0.0787 to 0.0630.
Although \textbf{w/o Audit} has the highest raw benefit density (1.7390), its weaker image-level and residual metrics show that high accumulated benefit alone does not guarantee a clean stroke allocation.
The isolated \textbf{Prune Only}, \textbf{Split Only}, and \textbf{Respawn Only} variants improve different aspects but do not maintain the same balance as the complete edit loop, indicating that the three operations are complementary.



\section{Conclusion}
We presented a model-guided optimization framework for quadratic Bezier-based SBR, featuring feedforward structural initialization and stroke-level quality auditing. Unlike fragmented polylines or handcrafted coarse-to-fine blur, our method uses compact continuous curves, a learned stroke-scale initializer, and test-time optimization to preserve object contours and spatial layout under fixed stroke budgets. The auditing mechanism evaluates each stroke via purity, benefit density, and residual demand, enabling adaptive removal, splitting, and respawning. Experiments show consistent gains in fidelity, perceptual quality, and stroke reliability, demonstrating that effective SBR requires not just differentiable optimization, but also structurally aware initialization and explicit local stroke evaluation.

\section{Acknowledgments}
This work was supported by the Science and Technology Commission of Shanghai Municipality under research grant No. 25ZR1401187, and it was financially supported by the Open Research Fund from Guangdong Laboratory of Artificial Intelligence and Digital Economy (SZ), under Grant NO.GML-KF-26-07. This work was also supported by the Shanghai Municipal Commission of Economy and Informatization under the Leading Industries Innovation and Development Program (Artificial Intelligence), Project No. 2026-GZL-RGZN-01020.

\clearpage
\bibliographystyle{plain}
\bibliography{main}

@inproceedings{haeberli1990paint,
  author = {Haeberli, Paul},
  title = {Paint by Numbers: Abstract Image Representations},
  booktitle = {Proceedings of SIGGRAPH},
  pages = {207--214},
  year = {1990},
  doi = {10.1145/97879.97902}
}

@inproceedings{meier1996painterly,
  author = {Meier, Barbara J.},
  title = {Painterly Rendering for Animation},
  booktitle = {Proceedings of SIGGRAPH},
  pages = {477--484},
  year = {1996},
  doi = {10.1145/237170.237288}
}

@inproceedings{litwinowicz1997processing,
  author = {Litwinowicz, Peter},
  title = {Processing Images and Video for an Impressionist Effect},
  booktitle = {Proceedings of SIGGRAPH},
  pages = {407--414},
  year = {1997},
  doi = {10.1145/258734.258893}
}

@inproceedings{curtis1997watercolor,
  author = {Curtis, Cassidy J. and Anderson, Sean E. and Seims, Joshua E. and Fleischer, Kurt W. and Salesin, David H.},
  title = {Computer-Generated Watercolor},
  booktitle = {Proceedings of SIGGRAPH},
  pages = {421--430},
  year = {1997},
  doi = {10.1145/258734.258896}
}

@inproceedings{hertzmann1998painterly,
  author = {Hertzmann, Aaron},
  title = {Painterly Rendering with Curved Brush Strokes of Multiple Sizes},
  booktitle = {Proceedings of SIGGRAPH},
  pages = {453--460},
  year = {1998},
  doi = {10.1145/280814.280951}
}

@inproceedings{hertzmann2001paint,
  author = {Hertzmann, Aaron},
  title = {Paint by Relaxation},
  booktitle = {Proceedings of Computer Graphics International},
  pages = {47--54},
  year = {2001},
  doi = {10.1109/CGI.2001.934657}
}

@article{decarlo2002stylization,
  author = {DeCarlo, Douglas and Santella, Anthony},
  title = {Stylization and Abstraction of Photographs},
  journal = {ACM Transactions on Graphics},
  volume = {21},
  number = {3},
  pages = {769--776},
  year = {2002},
  doi = {10.1145/566654.566650}
}

@article{winnemoller2006realtime,
  author = {Winnem{\"o}ller, Holger and Olsen, Sven C. and Gooch, Bruce},
  title = {Real-Time Video Abstraction},
  journal = {ACM Transactions on Graphics},
  volume = {25},
  number = {3},
  pages = {1221--1226},
  year = {2006},
  doi = {10.1145/1141911.1142018}
}

@article{kang2009flow,
  author = {Kang, Henry and Lee, Seungyong and Chui, Charles K.},
  title = {Flow-Based Image Abstraction},
  journal = {IEEE Transactions on Visualization and Computer Graphics},
  volume = {15},
  number = {1},
  pages = {62--76},
  year = {2009},
  doi = {10.1109/TVCG.2008.81}
}

@article{kyprianidis2013state,
  author = {Kyprianidis, Jan Eric and Collomosse, John P. and Wang, Tinghuai and Isenberg, Tobias},
  title = {State of the ``Art'': A Taxonomy of Artistic Stylization Techniques for Images and Video},
  journal = {IEEE Transactions on Visualization and Computer Graphics},
  volume = {19},
  number = {5},
  pages = {866--885},
  year = {2013},
  doi = {10.1109/TVCG.2012.160}
}

@inproceedings{ha2018sketchrnn,
  author = {Ha, David and Eck, Douglas},
  title = {A Neural Representation of Sketch Drawings},
  booktitle = {Proceedings of ICLR},
  year = {2018},
  url = {https://openreview.net/forum?id=Hy6GHpkCW}
}

@inproceedings{ganin2018spiral,
  author = {Ganin, Yaroslav and Kulkarni, Tejas and Babuschkin, Igor and Eslami, S. M. Ali and Vinyals, Oriol},
  title = {Synthesizing Programs for Images Using Reinforced Adversarial Learning},
  booktitle = {Proceedings of ICML},
  pages = {1652--1661},
  year = {2018},
  url = {http://proceedings.mlr.press/v80/ganin18a.html}
}

@article{nolte2023survey,
  author = {Nolte, Florian and Melnik, Andrew and Ritter, Helge J.},
  title = {Stroke-Based Rendering: From Heuristics to Deep Learning},
  journal = {CoRR},
  volume = {abs/2302.00595},
  year = {2023},
  doi = {10.48550/arXiv.2302.00595}
}

@article{tang2025attentionpainter,
  author = {Tang, Yizhe and Wang, Yue and Hu, Teng and Yi, Ran and Tan, Xin and Ma, Lizhuang and Lai, Yu-Kun and Rosin, Paul L.},
  title = {AttentionPainter: An Efficient and Adaptive Stroke Predictor for Scene Painting},
  journal = {IEEE Transactions on Visualization and Computer Graphics},
  pages = {10897--10911},
  year = {2025},
  doi = {10.1109/TVCG.2025.3618184}
}

@article{li2020diffvg,
  author = {Li, Tzu-Mao and Luk{\'a}{\v{c}}, Michal and Gharbi, Micha{\"e}l and Ragan-Kelley, Jonathan},
  title = {Differentiable Vector Graphics Rasterization for Editing and Learning},
  journal = {ACM Transactions on Graphics},
  volume = {39},
  number = {6},
  pages = {193:1--193:15},
  year = {2020},
  doi = {10.1145/3414685.3417871}
}

@inproceedings{egiazarian2020deep,
  author = {Egiazarian, Vage and Voynov, Oleg and Artemov, Alexey and Volkhonskiy, Denis and Safin, Aleksandr and Taktasheva, Maria and Zorin, Denis and Burnaev, Evgeny},
  title = {Deep Vectorization of Technical Drawings},
  booktitle = {Proceedings of ECCV},
  pages = {582--598},
  year = {2020},
  doi = {10.1007/978-3-030-58601-0_35}
}

@inproceedings{carlier2020deepsvg,
  author = {Carlier, Alexandre and Danelljan, Martin and Alahi, Alexandre and Timofte, Radu},
  title = {DeepSVG: A Hierarchical Generative Network for Vector Graphics Animation},
  booktitle = {Advances in Neural Information Processing Systems},
  year = {2020},
  url = {https://proceedings.neurips.cc/paper/2020/hash/bcf9d6bd14a2095866ce8c950b702341-Abstract.html}
}

@inproceedings{reddy2021im2vec,
  author = {Reddy, Pradyumna},
  title = {Im2Vec: Synthesizing Vector Graphics Without Vector Supervision},
  booktitle = {Proceedings of CVPR Workshops},
  pages = {2124--2133},
  year = {2021},
  doi = {10.1109/CVPRW53098.2021.00241}
}

@inproceedings{frans2022clipdraw,
  author = {Frans, Kevin and Soros, Lisa B. and Witkowski, Olaf},
  title = {CLIPDraw: Exploring Text-to-Drawing Synthesis through Language-Image Encoders},
  booktitle = {Advances in Neural Information Processing Systems},
  year = {2022},
  url = {https://proceedings.neurips.cc/paper_files/paper/2022/hash/21f76686538a5f06dc431efea5f475f5-Abstract-Conference.html}
}

@inproceedings{jain2023vectorfusion,
  author = {Jain, Ajay and Xie, Amber and Abbeel, Pieter},
  title = {VectorFusion: Text-to-SVG by Abstracting Pixel-Based Diffusion Models},
  booktitle = {Proceedings of CVPR},
  pages = {1911--1920},
  year = {2023},
  doi = {10.1109/CVPR52729.2023.00190}
}

@inproceedings{xing2023diffsketcher,
  author = {Xing, Ximing and Wang, Chuang and Zhou, Haitao and Zhang, Jing and Yu, Qian and Xu, Dong},
  title = {DiffSketcher: Text Guided Vector Sketch Synthesis through Latent Diffusion Models},
  booktitle = {Advances in Neural Information Processing Systems},
  year = {2023},
  url = {https://proceedings.neurips.cc/paper_files/paper/2023/hash/333e67fc4728f147d31608db3ca78e09-Abstract-Conference.html}
}

@inproceedings{chen2024vectorization,
  author = {Chen, Ye and Ni, Bingbing and Liu, Jinfan and Huang, Xiaoyang and Chen, Xuanhong},
  title = {Towards High-Fidelity Artistic Image Vectorization via Texture-Encapsulated Shape Parameterization},
  booktitle = {Proceedings of CVPR},
  pages = {15877--15886},
  year = {2024},
  doi = {10.1109/CVPR52733.2024.01503}
}

@article{liu2025bezier,
  author = {Liu, Xi and Zhou, Chaoyi and Zhao, Nanxuan and Huang, Siyu},
  title = {B{\'e}zier Splatting for Fast and Differentiable Vector Graphics},
  journal = {CoRR},
  volume = {abs/2503.16424},
  year = {2025},
  doi = {10.48550/arXiv.2503.16424}
}

@article{canny1986edge,
  author = {Canny, John F.},
  title = {A Computational Approach to Edge Detection},
  journal = {IEEE Transactions on Pattern Analysis and Machine Intelligence},
  volume = {8},
  number = {6},
  pages = {679--698},
  year = {1986},
  doi = {10.1109/TPAMI.1986.4767851}
}

@article{kass1988snakes,
  author = {Kass, Michael and Witkin, Andrew P. and Terzopoulos, Demetri},
  title = {Snakes: Active Contour Models},
  journal = {International Journal of Computer Vision},
  volume = {1},
  number = {4},
  pages = {321--331},
  year = {1988},
  doi = {10.1007/BF00133570}
}

@article{itti1998saliency,
  author = {Itti, Laurent and Koch, Christof and Niebur, Ernst},
  title = {A Model of Saliency-Based Visual Attention for Rapid Scene Analysis},
  journal = {IEEE Transactions on Pattern Analysis and Machine Intelligence},
  volume = {20},
  number = {11},
  pages = {1254--1259},
  year = {1998},
  doi = {10.1109/34.730558}
}

@article{wang2004ssim,
  author = {Wang, Zhou and Bovik, Alan C. and Sheikh, Hamid R. and Simoncelli, Eero P.},
  title = {Image Quality Assessment: From Error Visibility to Structural Similarity},
  journal = {IEEE Transactions on Image Processing},
  volume = {13},
  number = {4},
  pages = {600--612},
  year = {2004},
  doi = {10.1109/TIP.2003.819861}
}

@inproceedings{zhang2018lpips,
  author = {Zhang, Richard and Isola, Phillip and Efros, Alexei A. and Shechtman, Eli and Wang, Oliver},
  title = {The Unreasonable Effectiveness of Deep Features as a Perceptual Metric},
  booktitle = {Proceedings of CVPR},
  pages = {586--595},
  year = {2018},
  doi = {10.1109/CVPR.2018.00068}
}

@inproceedings{im2oil,
  title={Im2oil: Stroke-based oil painting rendering with linearly controllable fineness via adaptive sampling},
  author={Tong, Zhengyan and Wang, Xiaohang and Yuan, Shengchao and Chen, Xuanhong and Wang, Junjie and Fang, Xiangzhong},
  booktitle={Proceedings of the 30th ACM international conference on multimedia},
  pages={1035--1046},
  year={2022}
}

@inproceedings{snpsd,
  title={Stroke-based neural painting and stylization with dynamically predicted painting region},
  author={Hu, Teng and Yi, Ran and Zhu, Haokun and Liu, Liang and Peng, Jinlong and Wang, Yabiao and Wang, Chengjie and Ma, Lizhuang},
  booktitle={Proceedings of the 31st ACM International Conference on Multimedia},
  pages={7470--7480},
  year={2023}
}

@inproceedings{snp,
  title={Stylized neural painting},
  author={Zou, Zhengxia and Shi, Tianyang and Qiu, Shuang and Yuan, Yi and Shi, Zhenwei},
  booktitle={Proceedings of the IEEE/CVF Conference on Computer Vision and Pattern Recognition},
  pages={15689--15698},
  year={2021}
}

@inproceedings{p-trans,
  title={Paint transformer: Feed forward neural painting with stroke prediction},
  author={Liu, Songhua and Lin, Tianwei and He, Dongliang and Li, Fu and Deng, Ruifeng and Li, Xin and Ding, Errui and Wang, Hao},
  booktitle={Proceedings of the IEEE/CVF international conference on computer vision},
  pages={6598--6607},
  year={2021}
}

@inproceedings{p-rl,
  title={Learning to paint with model-based deep reinforcement learning},
  author={Huang, Zhewei and Heng, Wen and Zhou, Shuchang},
  booktitle={Proceedings of the IEEE/CVF international conference on computer vision},
  pages={8709--8718},
  year={2019}
}

@article{kerbl20233dgs,
  author = {Kerbl, Bernhard and Kopanas, Georgios and Leimk{\"u}hler, Thomas and Drettakis, George},
  title = {3{D} {G}aussian Splatting for Real-Time Radiance Field Rendering},
  journal = {ACM Transactions on Graphics},
  volume = {42},
  number = {4},
  pages = {139:1--139:14},
  year = {2023},
  doi = {10.1145/3592433}
}

@article{liu2026dualstroke,
  author = {Liu, Jinfan and Zhang, Wuze and Hu, Zhangli and Zhao, Zhehan and Chen, Ye and Ni, Bingbing},
  title = {Differentiable Stroke Planning with Dual Parameterization for Efficient and High-Fidelity Painting Creation},
  journal = {CoRR},
  volume = {abs/2604.02752},
  year = {2026},
  doi = {10.48550/arXiv.2604.02752}
}

@InProceedings{div2k,
	author = {Agustsson, Eirikur and Timofte, Radu},
	title = {NTIRE 2017 Challenge on Single Image Super-Resolution: Dataset and Study},
	booktitle = {The IEEE Conference on Computer Vision and Pattern Recognition (CVPR) Workshops},
	month = {July},
	year = {2017}
}

@inproceedings{segpainting2024,
  author    = {Manuel Ladron de Guevara and Matt Fisher and Aaron Hertzmann},
  title     = {Segmentation-Based Parametric Painting},
  booktitle = {2024 IEEE International Conference on Multimedia and Expo Workshops (ICMEW)},
  pages     = {1--6},
  year      = {2024},
  doi       = {10.1109/ICMEW63481.2024.10645426}
}

@inproceedings{hu2024artist,
  author    = {Zhangli Hu and Ye Chen and Zhongyin Zhao and Jinfan Liu and Bilian Ke and Bingbing Ni},
  title     = {Towards Artist-Like Painting Agents with Multi-Granularity Semantic Alignment},
  booktitle = {Proceedings of the 32nd ACM International Conference on Multimedia (MM '24)},
  pages     = {10191--10199},
  year      = {2024},
  doi       = {10.1145/3664647.3681657}
}

\newpage

\appendix
\section{Supplementary Material}

\subsection{Analytic Distance to Quadratic Bezier}

\label{sec:analytic_distance_quadratic_bezier}
Consider a pixel center $\mathbf{p}\in\mathbb{R}^2$ and a quadratic Bezier curve defined by control points $P_0, P_1, P_2 \in \mathbb{R}^2$:
\begin{equation}
\mathbf{B}(t) = (1-t)^2 P_0 + 2(1-t)t P_1 + t^2 P_2, \quad t\in[0,1].
\end{equation}
The point-to-curve distance is defined as
\begin{equation}
D(\mathbf{p}) = \min_{t\in[0,1]} \|\mathbf{B}(t) - \mathbf{p}\|_2 .
\end{equation}
To determine the minimizer, we minimize the squared distance. Let
\begin{equation}
u = P_0 - 2P_1 + P_2, \quad v = P_1 - P_0, \quad w = P_0 - \mathbf{p}.
\end{equation}
The curve can then be rewritten as
\begin{equation}
\mathbf{B}(t) = P_0 + 2vt + ut^2 .
\end{equation}

\subsubsection*{Non-Degenerate Case}

For non-degenerate quadratic curves with $\|u\|_2^2 > \epsilon$, setting the derivative of $\|\mathbf{B}(t)-\mathbf{p}\|_2^2$ to zero yields the cubic equation
\begin{equation}
(u^\top u)\,t^3 + 3(u^\top v)\,t^2 + (2v^\top v + u^\top w)\,t + v^\top w = 0 .
\end{equation}
Denoting the cubic as $a t^3 + b t^2 + c t + d = 0$, we apply the substitution $t = x - b/(3a)$ to eliminate the quadratic term. With
\begin{equation}
t = x - \frac{u^\top v}{u^\top u},
\end{equation}
we obtain the depressed cubic
\begin{equation}
x^3 + px + q = 0 .
\end{equation}
Its root structure is governed by the discriminant
\begin{equation}
\Delta = \left(\frac{q}{2}\right)^2 + \left(\frac{p}{3}\right)^3 .
\end{equation}

\paragraph{Case $\Delta > 0$.}
There is one real root and two complex conjugates. The real root is given by
\begin{equation}
x = \operatorname{cbrt}\!\left(-\frac{q}{2} + \sqrt{\Delta}\right) + \operatorname{cbrt}\!\left(-\frac{q}{2} - \sqrt{\Delta}\right),
\end{equation}
where $\operatorname{cbrt}(\cdot)$ denotes the real cube root.

\paragraph{Case $\Delta \leq 0$.}
The cubic has three real roots (counting multiplicity). If $p=q=0$, the triple root is $x=0$, giving $t=-b/(3a)$; this case is handled directly without evaluating the trigonometric expression. Otherwise, we employ the trigonometric form
\begin{equation}
x_k = 2\sqrt{-\frac{p}{3}}\,\cos\!\left(\frac{\theta + 2\pi k}{3}\right), \quad k = 0,1,2,
\end{equation}
with
\begin{equation}
\theta = \arccos\!\left( \operatorname{clamp}\!\left( -\frac{q}{2} \Big/ \sqrt{-\left(\frac{p}{3}\right)^3},\; -1,\; 1 \right) \right).
\end{equation}
Repeated roots are removed numerically to avoid redundant candidates.

\subsubsection*{Candidate Selection}

All real roots falling within the interval $[0,1]$ are collected. Together with the endpoints $t=0$ and $t=1$, they form the candidate set $\mathcal{C}(\mathbf{p})$. The distance is then computed as
\begin{equation}
D(\mathbf{p}) = \min_{t \in \mathcal{C}(\mathbf{p})} \|\mathbf{B}(t) - \mathbf{p}\|_2 .
\end{equation}

\subsubsection*{Degenerate Cases}

When $\|u\|_2^2 \leq \epsilon$, the curve degenerates to a line segment. The distance reduces to the standard point-to-segment distance: the projection of $\mathbf{p}$ onto the segment $[P_0, P_2]$ is computed and the projection parameter is clamped to $[0,1]$. If additionally $\|P_2 - P_0\|_2^2 \leq \epsilon$, the segment collapses to a point and the distance becomes $\|\mathbf{p} - P_0\|_2$. Throughout all experiments, coordinates are normalized to $[0,1]^2$ and we set $\epsilon = 10^{-6}$.

\subsubsection*{Differentiability}

This formulation yields the exact closest-point distance up to numerical precision. During optimization, we first determine the closest parameter $t^*$ and then back-propagate through $\|\mathbf{B}(t^*) - \mathbf{p}\|_2$, treating the discrete candidate selection as fixed. The distance is differentiable almost everywhere, with non-differentiable points occurring only when multiple candidates attain the exact same minimum or when the closest candidate switches—both constitute measure-zero sets.

\subsection{Detailed Model Architecture}
\label{sec:detailed_model_architecture}
The feedforward predictor operates patch-wise. A full target image \(I_{gt}\in\mathbb{R}^{H\times W\times 3}\) is divided into non-overlapping \(256\times256\) patches \(\{x_k\}_{k=1}^{K}\). Each patch is processed independently, and the resulting local stroke sets \(\mathcal{S}_k\) are transformed to global coordinates and concatenated as \(\mathcal{S}=\bigcup_{k=1}^{K}\mathcal{S}_k\). Both training and inference use non-overlapping patches. After coordinate conversion, all strokes are composited on a shared global canvas. Since the stroke widths are non-zero, the stroke support can extend beyond the original patch boundaries, and the subsequent global optimization jointly refines the strokes across the full canvas.  We observe no visible checkerboard artifacts at patch seams.

\paragraph{Backbone.}
Each \(256\times256\) patch is resized to \(224\times224\), normalized with ImageNet statistics, and encoded by \texttt{vit\_small\_patch16\_224} (from \texttt{timm}). The CLS token is discarded, retaining \(M=196\) spatial patch tokens, which are linearly projected to dimension \(d\): \(F_k = W_f\,\mathrm{Backbone}(x_k)\in\mathbb{R}^{M\times d}\).

\begin{figure*}[t]
    \centering
    \includegraphics[width=\linewidth]{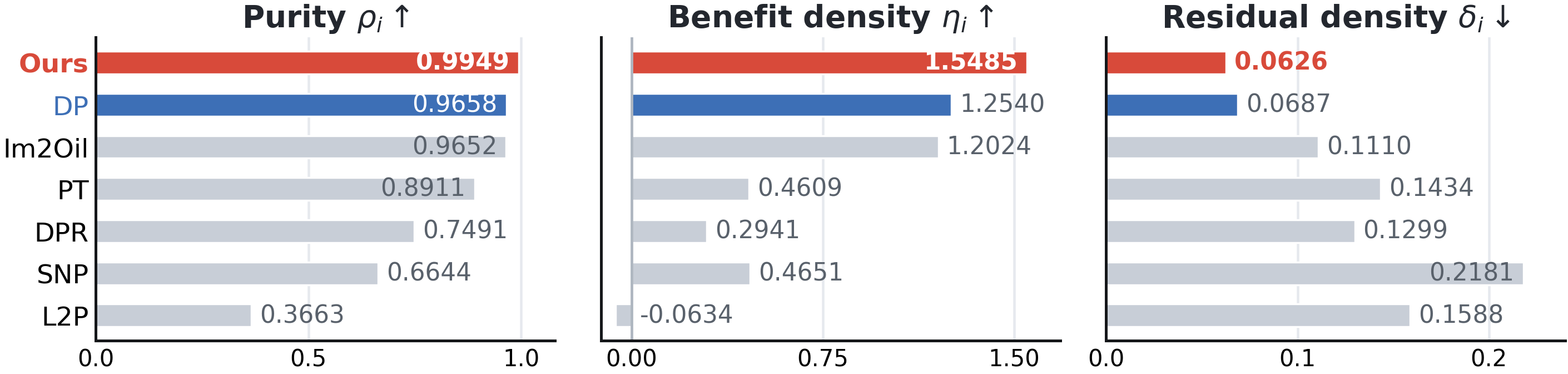}
    \caption{
    \textbf{Stroke quality comparison.}
    Comparison of stroke auditing metrics on DIV2K($600\times600$). Our framework establishes a significant margin over all baselines, demonstrating that it generates locally constructive and structurally meaningful strokes rather than unstructured pixel accumulations.
    }
    \label{fig:stroke_quality}
\end{figure*}

\paragraph{Template initialization.}
The \(N\) strokes per patch are partitioned into coarse, mid, and fine groups (default ratio 20\%/30\%/50\%). Each group uses stage-specific geometric priors summarized in Table~\ref{tab:template_priors}. Given a center $C$ uniformly sampled within the patch, tangent direction \(\mathbf{t}\), length \(\ell\), and curvature offset \(\kappa\), control points are initialized as \(P_0=C-\frac{\ell}{2}\mathbf{t}\), \(P_1=C+\kappa\,\mathbf{n}\), \(P_2=C+\frac{\ell}{2}\mathbf{t}\), where \(\mathbf{n}\) is the unit normal obtained by rotating \(\mathbf{t}\) counterclockwise by \(90^\circ\). The stroke color is initialized by directly sampling the RGB values from the target image $I_{gt}$ at the center coordinate $C$, while the initial opacity (alpha) is fixed at $0.7$ for all strokes across all scales.
The initial queries sum learnable group embeddings and sinusoidal stroke-index encodings. Decoded residuals are added to the geometric templates, as defined in the main method.

\begin{table}[t]
\centering
\caption{Stage-specific geometric priors for stroke templates normalized to local patch coordinates.}
\renewcommand{\arraystretch}{0.95} 
\setlength{\tabcolsep}{3.2pt}    
\label{tab:template_priors}
\resizebox{0.92\linewidth}{!}{
\begin{tabular}{lcccp{4cm}}
\toprule
\textbf{Group} & \textbf{Ratio} & \textbf{Curve length} & \textbf{Width} & \textbf{Role} \\
\midrule
Coarse & 20\% & 0.24--0.56 & 0.12--0.30 & large color regions \\
Mid    & 30\% & 0.08--0.22 & 0.04--0.12 & medium-scale boundaries \\
Fine   & 50\% & 0.008--0.06 & 0.006--0.03 & local residual hints \\
\bottomrule
\end{tabular}}
\end{table}

\paragraph{Decoder.}
The decoder is organized into three cascaded stages, each implemented as a lightweight transformer decoder layer with 6 heads, model dimension 384, feedforward expansion ratio 4, and pre-LayerNorm.
For stage-wise feature aggregation, keys and values are image features for the coarse stage, image features concatenated with coarse outputs for the mid stage, and image features concatenated with both coarse and mid outputs for the fine stage; attention among stroke queries is restricted within each group.
Prediction heads share a 2-hidden-layer MLP of dimension 256 with GELU activation and scaling factor 0.5. 
Crucially, the combination of the stage-specific priors (Table~\ref{tab:template_priors}) and the strictly bounded residual outputs ($\gamma \tanh$) acts as an implicit size constraint. This prevents fine strokes from exceeding their intended local receptive fields and collapsing into the coarse scale during end-to-end training.
After decoding, normalized stroke coordinates are scaled to \(256\times256\) patch resolution and offset by the patch location to obtain global stroke coordinates.

\paragraph{Depth and compositing order.}
Depth determines the compositing order and is not a continuously optimized stroke parameter. Each stroke receives a fixed depth key at birth, with stroke indices resolving ties. Surviving strokes retain their depth keys during optimization; split children inherit their parent's depth, and newly spawned strokes receive fixed birth-time assignments. To assess sensitivity to this fixed ordering, Sec.~\ref{sec:fixed_order_ablation} compares alternative rendering orders within three fixed depth intervals while preserving the order between intervals.

\paragraph{Training Details.}The network is trained end-to-end on Flickr2K using the AdamW optimizer with a base learning rate of $1 \times 10^{-4}$, weight decay of $1 \times 10^{-4}$, and a total batch size of 32. We employ a cosine annealing learning rate scheduler with a linear warmup over the first 5\% of total iterations. The loss weights balancing the reconstruction and structural constraints are empirically set to $\lambda_{\ell_2} = 1.0$ and $\lambda_{\text{edge}} = 0.5$. The model is trained for 200 epochs, about 8 hours, on a single compute node equipped with 4 NVIDIA RTX 4090 GPUs. Since the parameter bounds and color initialization provide a strong starting point, the network converges stably without requiring complex multi-stage curriculum learning.

\paragraph{Computational Complexity.}A major advantage of our feedforward architecture is the efficiency of the structural initialization stage: it produces an organized stroke layout in a single forward pass, instead of relying on slow per-image optimization or search to discover the layout. Note that this claim concerns the initialization stage only; the full pipeline still applies test-time optimization for high-fidelity refinement (Sec.~\ref{sec:exp}). For comparison, gradient-based optimization from random initialization requires over 600 iterations ($\sim$3s) to match the PSNR of the native feedforward output (Table~\ref{tab:ablation}), about 20$\times$ slower than the 0.15s initializer. The model comprises approximately 28M parameters (ViT-small backbone and the cascaded decoder). For a high-resolution target image of size $1024 \times 1024$, the image is divided into 16 non-overlapping $256 \times 256$ patches. Because our patch-wise design contains no inter-patch dependencies, these 16 patches are processed in parallel during the forward pass. On a single NVIDIA RTX 4090 GPU, the initialization pipeline---including token extraction, multi-scale stroke prediction, and the final B\'{e}zier splatting rendering---takes approximately 0.15 seconds ($\sim$6 FPS) per $1024 \times 1024$ image, making it highly suitable for practical downstream applications.

\begin{figure*}[t]
\centering
\includegraphics[width=\linewidth]{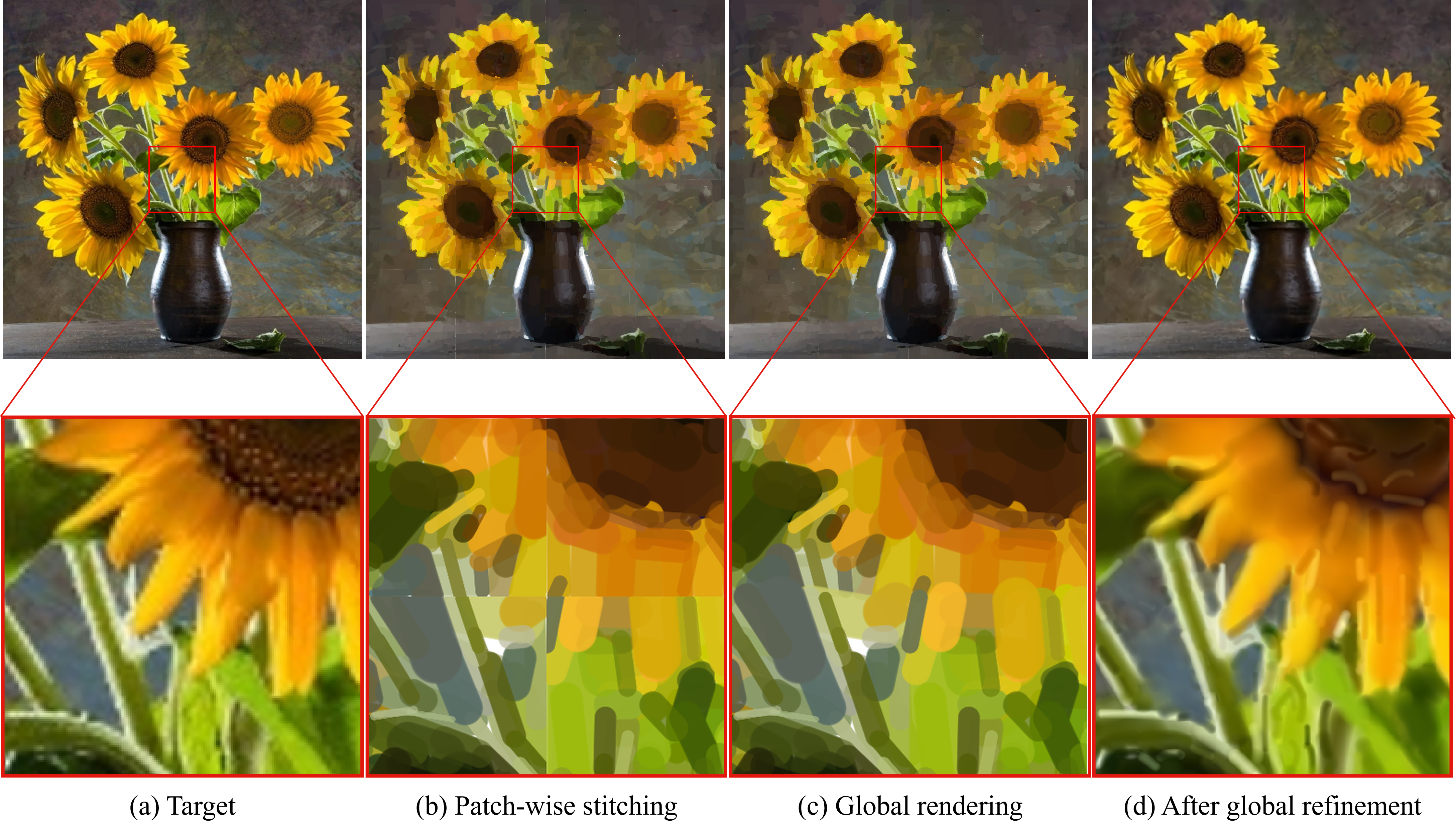}
\caption{\textbf{Patch-boundary behavior before and after global refinement.} Columns show the target, patch-wise stitching, global rendering of the same initialized strokes, and the result after 5,000 global optimization steps. The bottom row enlarges the patch junction. Patches are $256\times256$ on a $1024\times1024$ base canvas.}
\label{fig:patch_boundary_behavior}
\end{figure*}

\paragraph{Cross-patch consistency.}
\label{sec:patch_boundary_behavior}
Predicted strokes are transformed from patch-local coordinates and composited on a shared canvas. Although initial control points are confined to their source patches, finite stroke widths allow rendered support to cross patch boundaries. Subsequent global optimization jointly refines all strokes and permits control points to move beyond their original patches. Figure~\ref{fig:patch_boundary_behavior} compares independent patch rendering followed by stitching, global rendering of the same initialization, and the globally refined result. The matched junction crops show that patch-correlated structure can remain after initialization and is substantially reduced by global refinement. This behavior supports cross-patch coherence without imposing an explicit boundary-continuity constraint.

\subsection{Direction Alignment Loss}
\label{sec:direction_alignment_loss}
To encourage coherent stroke geometry, we regularize each quadratic stroke against a target image direction field. We first estimate a multi-scale direction field from the target image using structure-tensor analysis at fine and coarse scales, and blend the two fields according to local coherence. Let $B_i(t)$ denote the $i$-th quadratic Bezier stroke. For each stroke, we sample three positions $t_m \in \{0.25, 0.5, 0.75\}$ and compute the corresponding unit tangent vectors $\tau_{i,m}$. At the same locations, we read the local unit direction vectors $d_{i,m}$ from the target direction field together with coherence scores $c_{i,m} \in [0,1]$.

The direction loss is defined as
\begin{equation}
\mathcal{L}_{\text{dir}}
=
\frac{1}{N}
\sum_{i=1}^{N}
\frac{
\sum_{m=1}^{3}
w_m\, c_{i,m}\,
\bigl(1 - (\tau_{i,m}^{\top} d_{i,m})^2\bigr)
}{
\sum_{m=1}^{3} w_m\, c_{i,m} + \epsilon
},
\end{equation}
where $N$ is the number of active strokes, $(w_1,w_2,w_3)=(0.25,0.5,0.25)$ are fixed sampling weights, and $\epsilon$ is a small constant for numerical stability. This formulation is orientation-invariant, since reversing the tangent direction does not change the squared inner product. The coherence weights suppress unreliable supervision in weakly structured or locally isotropic regions, so the loss mainly acts where the target image provides a clear directional cue.

\subsection{Additional Ablation Experiments}
\label{sec:additional_ablations}

\subsubsection{Fixed Within-Layer Ordering}
\label{sec:fixed_order_ablation}
\begin{figure*}[t]
\centering
\includegraphics[width=\linewidth]{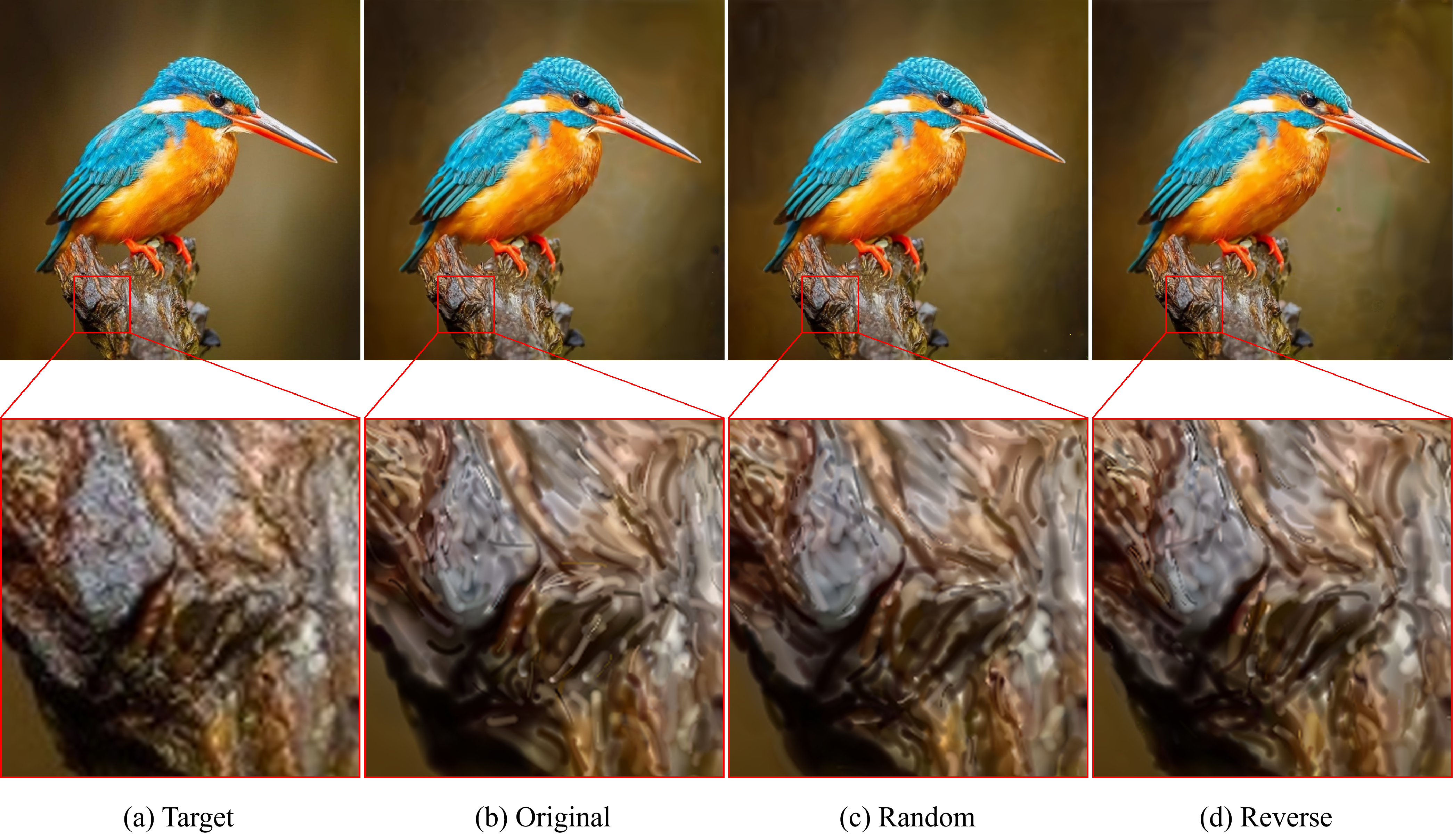}
\caption{\textbf{Sensitivity to fixed within-layer ordering.} Columns show the target and results under original, random, and reversed within-layer orders. Full images appear above, with the same region enlarged below. All conditions start from identical stroke geometry, colors, and opacities, differing only in rendering order, and use the same optimization budget. Within-layer reordering has little effect on reconstruction quality.}
\label{fig:fixed_order_ablation}
\end{figure*}

We compare original, random, and reversed within-layer orders while preserving layer membership and between-layer ordering. Each condition starts from identical non-depth parameters and undergoes 5,000 optimization steps; depth keys remain fixed. The diagnostic uses three numerical depth strata and the same $600\times600$, 5,000-stroke setting throughout. Figure~\ref{fig:fixed_order_ablation} shows full reconstructions and matched overlap details, while Table~\ref{tab:gallery8_order} summarizes the gallery results. Using the same reconstruction-based snapshot selection as the main pipeline, the mean PSNR differs by approximately 0.07 dB across the tested orders, with similar SSIM and LPIPS. These aggregate results support the practical use of fixed within-layer ordering.

\begin{table}[ht]
\centering
\caption{\textbf{Within-layer stroke-order ablation.} Mean results on Im2Oil Gallery under the $600\times600$, 5,000-stroke setting. Every condition uses the reconstruction-selected output exported by the pipeline.}
\label{tab:gallery8_order}
\resizebox{0.6\linewidth}{!}{%
\begin{tabular}{lccc}
\toprule
Order & PSNR $\uparrow$ & SSIM $\uparrow$ & LPIPS $\downarrow$ \\
\midrule
Original & 31.28 & 0.8704 & 0.1731 \\
Random & 31.35 & 0.8709 & 0.1702 \\
Reversed & 31.35 & 0.8697 & 0.1748 \\
\bottomrule
\end{tabular}
}
\end{table}

\subsubsection{Line-Segment Primitives}
\label{sec:line_primitive_ablation}
\begin{figure*}[t]
\centering
\includegraphics[width=\linewidth]{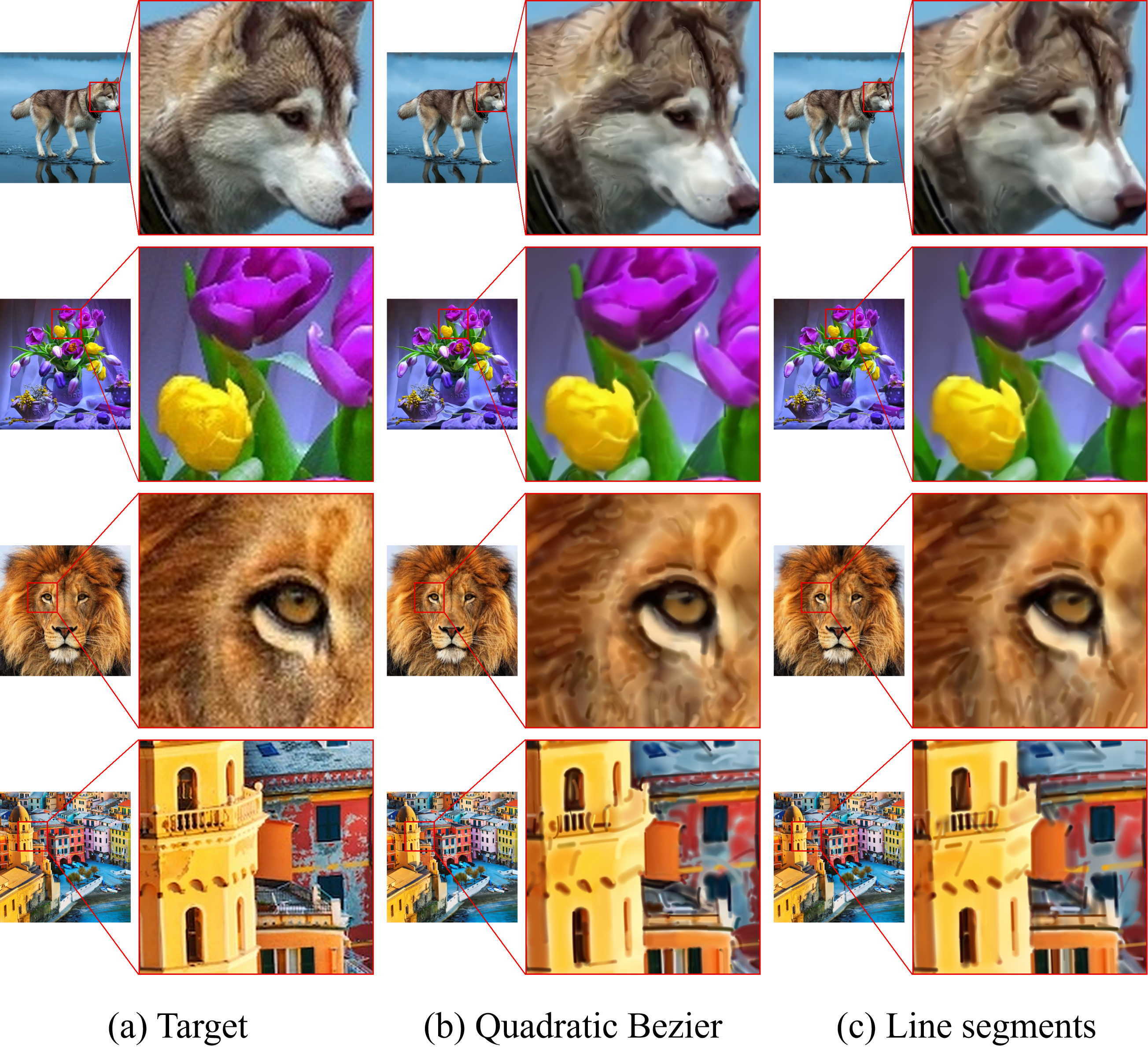}
\caption{\textbf{Quadratic B\'{e}zier and line-segment primitives.} Columns show the target, quadratic-stroke reconstruction, and line-segment reconstruction. Both variants use the same stroke capacity and optimization budget.}
\label{fig:line_primitive_ablation}
\end{figure*}

We replace quadratic curves with finite-width line segments and optimize their endpoints, widths, colors, and opacities using an analytic point-to-segment renderer. Both variants use the same initialization checkpoint, stroke capacity, optimization budget, and reconstruction-based snapshot selection. On Im2Oil Gallery under the $600\times600$, 5,000-stroke setting, quadratic strokes achieve mean PSNR/SSIM/LPIPS of 31.23 dB/0.8715/0.1707, compared with 30.14 dB/0.8507/0.1971 for line segments. Figure~\ref{fig:line_primitive_ablation} provides the corresponding visual comparison.

\subsection{Material Rendering and Enlarged Qualitative Results}
\label{sec:material_qualitative}
\begin{figure*}[!p]
\centering
\includegraphics[width=\linewidth]{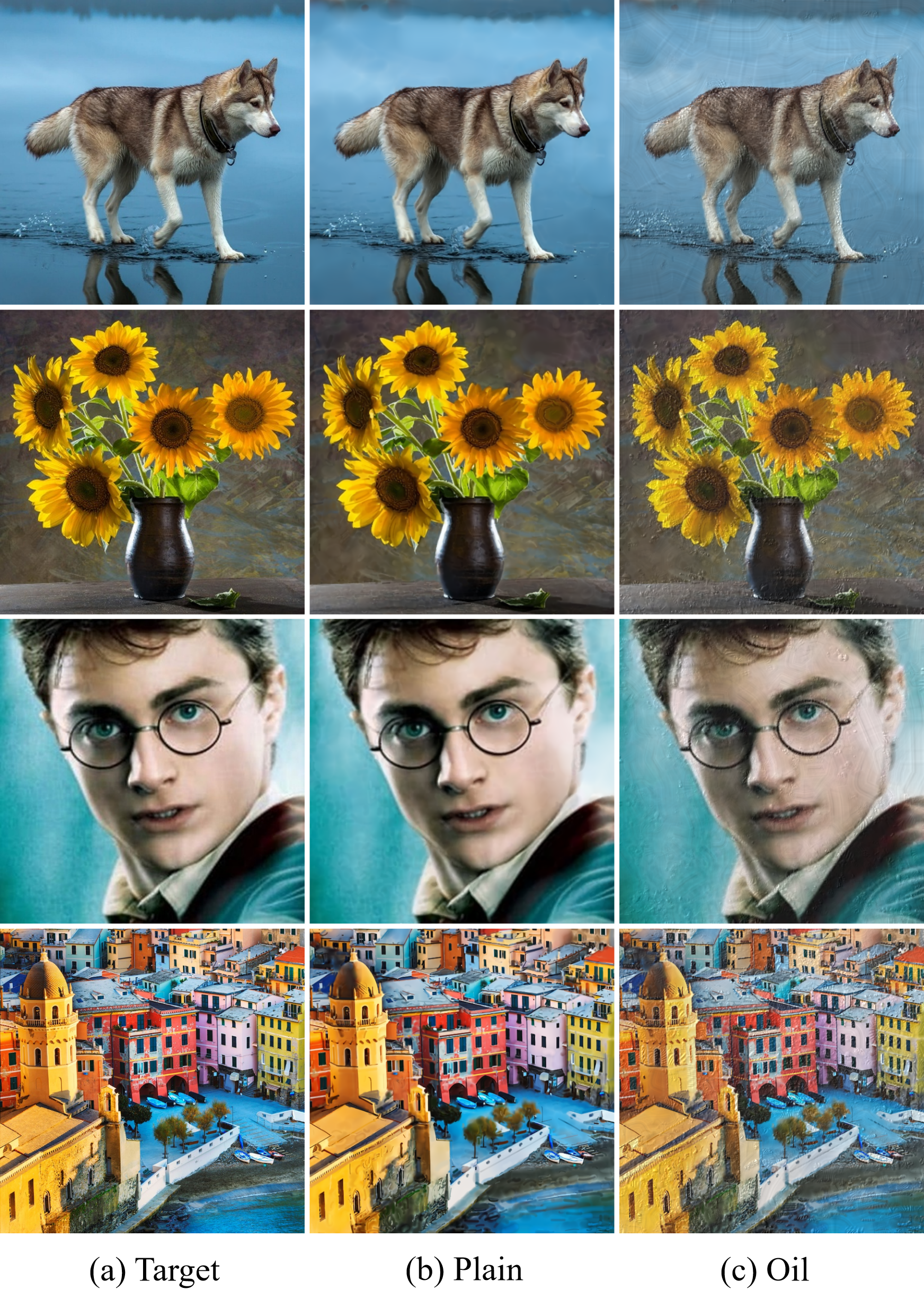}
\caption{\textbf{Enlarged qualitative results and oil rendering.} Columns show the input, plain rendering, and oil rendering. Both rendering variants share identical fitted stroke geometry and colors; enlarged outputs are rerasterizations without additional fitting.}
\label{fig:material_qualitative}
\end{figure*}

Figure~\ref{fig:material_qualitative} compares inputs, plain stroke renderings, and oil renderings. The plain and oil variants share the same fitted geometry and colors; brush and canvas textures, paint height, and lighting determine the material appearance without additional stroke optimization. For enlarged inspection, we rerasterize the fitted primitives at $2400\times2400$ from fits obtained on a $600\times600$ base canvas with $2\times$ supersampling.

\subsection{Dataset Licenses and Usage}
We use existing datasets only for academic research and do not redistribute the original images. Table~\ref{tab:dataset_licenses} reports the license or usage terms available from the official dataset or project releases.

\begin{table}[t]
\centering
\footnotesize
\caption{Dataset licenses or usage terms for the assets used in this paper.}
\label{tab:dataset_licenses}
\resizebox{\linewidth}{!}{
\begin{tabular}{p{2.4cm}p{2.1cm}p{6.5cm}p{4.2cm}}
\toprule
\textbf{Dataset} & \textbf{Use} & \textbf{License / terms} & \textbf{Source} \\
\midrule
Flickr2K & Training & No unified named dataset license is specified in the original release; images were collected from Flickr. We use the data for non-commercial academic training only and do not redistribute the images. & EDSR/Flickr2K release: \url{https://github.com/LimBee/NTIRE2017} \\
DIV2K~\cite{div2k} & Evaluation & Official usage terms restrict the dataset to academic research purposes; image copyrights remain with the original owners. & Official DIV2K page: \url{https://data.vision.ee.ethz.ch/cvl/DIV2K/} \\
Im2Oil Gallery~\cite{im2oil} & Evaluation & The official Im2Oil project repository is released under the CC0-1.0 license. & Official repository: \url{https://github.com/TZYSJTU/Im2Oil} \\
\bottomrule
\end{tabular}}
\end{table}

\subsection{User Study}
To evaluate perceptual painting quality beyond pixel-level metrics, we conducted a user study involving 100 participants. In each trial, a participant was shown a source image alongside seven corresponding painting results, one generated by our method and six by baseline approaches, presented in randomized order. Participants rated each result on a 1–5 Likert scale according to four painting-oriented criteria:
(a) \emph{Structural Planning}, which assesses whether strokes follow meaningful object structures and form a coherent global organization;
(b) \emph{Artistic Quality}, which captures the overall aesthetic appeal and painting-like appearance;
(c) \emph{Human-likeness}, which measures how closely the stroke arrangement resembles human painting behavior; and
(d) \emph{Stroke Quality}, which evaluates whether individual strokes are locally meaningful, coherent, and well aligned with image structures.
Table~\ref{tab:userstudy} reports the ratings averaged across all participants and test images. Our method achieves the highest scores on all four criteria, demonstrating that the proposed structural initialization and adaptive stroke auditing improve not only the final visual quality but also the perceived organization and quality of individual strokes.

\paragraph{Ethics, consent, and instructions.}
The user study was conducted as an anonymous, minimal-risk perceptual survey under the authors' institutional requirements for human-subject research. The study protocol was approved by the authors' institutional review process for anonymous perceptual surveys. Before participation, each participant was informed that the study was voluntary, that only anonymous rating data would be collected, that no personally identifiable information or sensitive attributes would be recorded, and that the ratings would be used only for academic research. All participants provided informed consent before taking the survey and could stop at any time. Participants were shown the following instruction: ``Please compare each painting result with the source image and independently rate the result from 1 to 5 according to structural planning, artistic quality, human-likeness, and stroke quality. There are no right or wrong answers; please rely on your visual judgment.'' No monetary compensation was provided.

\begin{table}[t]
\centering
\renewcommand{\arraystretch}{0.95} 
\setlength{\tabcolsep}{3.2pt}    
\resizebox{\linewidth}{!}{
\begin{tabular}{lcccc}
\toprule
\textbf{Method} &
\textbf{Struct. Plan.↑} &
\textbf{Artistic Qual.↑} &
\textbf{Human-like↑} &
\textbf{Stroke Qual.↑} \\
\midrule
SNP~\cite{snp} & 2.41 & 2.38 & 2.29 & 2.35 \\
PT~\cite{p-trans} & 2.78 & 2.85 & 2.71 & 2.74 \\
Im2Oil~\cite{im2oil} & 3.15 & 3.22 & 3.08 & 3.11 \\
DPR~\cite{snpsd} & 3.33 & 3.19 & 3.14 & 3.08 \\
L2P~\cite{p-rl} & 3.52 & 3.58 & 3.47 & 3.44 \\
DP~\cite{liu2026dualstroke} & 3.91 & 3.97 & 3.85 & 3.89 \\
\midrule
\textbf{Ours} & \textbf{4.31} & \textbf{4.39} & \textbf{4.24} & \textbf{4.36} \\
\bottomrule
\end{tabular}}
\caption{User study results on perceptual painting quality. Participants rated each result on a 1--5 Likert scale; higher scores indicate better perceptual quality.}
\label{tab:userstudy}
\end{table}

\subsection{Broader Impacts}
\label{sec:broader_impacts}
This work may have positive societal impacts in creative assistance and art education. First, it can automatically convert ordinary photographs into stroke-based artistic renderings, lowering the barrier for users to create painterly visual content. Second, the smooth quadratic Bezier stroke representation provides explicit and interpretable curve geometry, which can help beginners study stroke direction, curvature, and layout when learning digital painting. Third, the proposed continuous stroke primitive and structural initialization framework may support future controllable generative models by providing a compact, editable, and structure-aware intermediate representation for painterly image generation.
We do not identify direct high-risk negative societal impacts from the proposed method, as it does not generate new semantic content, release a high-risk model, or involve sensitive decision-making. As with general image stylization tools, users should apply the method only to images they have the right to process and should avoid using stylized outputs in contexts that misrepresent authorship or consent.

\subsection{Limitation and Future Work}
We present Quadratic Bezier Splatting, a model-guided optimization framework for stroke-based rendering that combines an analytic differentiable stroke representation with feedforward structural initialization and stroke-level quality auditing. Experiments show it achieves more compact and coherent layouts, better reconstruction and perceptual quality, and lower redundancy and optimization time than existing methods. However, limitations remain: (a) the auditing mechanism is mainly driven by reconstruction-oriented criteria such as stroke usefulness, purity, and residual demand, and may miss higher-level artistic factors such as intentional abstraction, brush rhythm, semantic expressiveness, or style-specific stroke ordering; (b) the semantic abstraction process behind feedforward structural initialization is not explicitly controllable, making it difficult to directly specify coarse-to-fine organization or assign precise semantic roles to individual strokes; (c) the current auditing strategy involves discrete stroke-level operations such as selecting, splitting, pruning, or refining strokes according to audit scores, which are not fully differentiable and cannot yet be seamlessly optimized end-to-end. Future work will explore: (a) style-aware and user-controllable auditing objectives; (b) interpretable and differentiable semantic controls for structural initialization; (c) differentiable approximations to stroke-level auditing and adjustment for unified end-to-end training.

\section*{NeurIPS Paper Checklist}

\begin{enumerate}

\item {\bf Claims}
    \item[] Question: Do the main claims made in the abstract and introduction accurately reflect the paper's contributions and scope?
    \item[] Answer: \answerYes{}.
    \item[] Justification: The abstract and Introduction state the three main contributions: quadratic B\'ezier splatting, feedforward structural initialization, and single-stroke auditing. The claims are supported by the method description in Methodology and by quantitative, qualitative, renderer, and ablation results in Sec.~\ref{sec:exp}.

\item {\bf Limitations}
    \item[] Question: Does the paper discuss the limitations of the work performed by the authors?
    \item[] Answer: \answerYes{}.
    \item[] Justification: The supplementary material includes a ``Limitation and Future Work'' subsection discussing reconstruction-oriented audit metrics that may miss higher-level artistic factors, limited explicit control over semantic stroke organization, and discrete auditing operations that are not fully differentiable.

\item {\bf Theory assumptions and proofs}
    \item[] Question: For each theoretical result, does the paper provide the full set of assumptions and a complete (and correct) proof?
    \item[] Answer: \answerYes{}.
    \item[] Justification: The paper's main analytical component is the exact point-to-curve distance used by quadratic B\'ezier splatting. Suppl.~\ref{sec:analytic_distance_quadratic_bezier} gives the assumptions, non-degenerate and degenerate cases, candidate selection, and differentiability discussion for this derivation.

    \item {\bf Experimental result reproducibility}
    \item[] Question: Does the paper fully disclose all the information needed to reproduce the main experimental results of the paper to the extent that it affects the main claims and/or conclusions of the paper (regardless of whether the code and data are provided or not)?
    \item[] Answer: \answerYes{}.
    \item[] Justification: Sec.~\ref{sec:exp} reports the evaluated datasets, resolutions, stroke budgets, baselines, metrics, model size, optimizer, training schedule, hardware, and inference/refinement setup. Additional architecture and training details are provided in Suppl.~\ref{sec:detailed_model_architecture}.

\item {\bf Open access to data and code}
    \item[] Question: Does the paper provide open access to the data and code, with sufficient instructions to faithfully reproduce the main experimental results, as described in supplemental material?
    \item[] Answer: \answerNo{}.
    \item[] Justification: A code package is not provided with this paper. The paper and supplementary material nevertheless provide detailed implementation and experimental information, including datasets, model architecture, loss terms, hyperparameters, training schedule, compute resources, evaluation settings, and ablation protocols.

\item {\bf Experimental setting/details}
    \item[] Question: Does the paper specify all the training and test details (e.g., data splits, hyperparameters, how they were chosen, type of optimizer) necessary to understand the results?
    \item[] Answer: \answerYes{}.
    \item[] Justification: Sec.~\ref{sec:exp} specifies datasets, image resolutions, stroke budgets, baselines, metrics, training optimizer, learning rate, weight decay, batch size, hardware, and test-time refinement steps. Suppl.~\ref{sec:detailed_model_architecture} further details the patch pipeline, ViT encoder, decoder, priors, losses, and training schedule.

\item {\bf Experiment statistical significance}
    \item[] Question: Does the paper report error bars suitably and correctly defined or other appropriate information about the statistical significance of the experiments?
    \item[] Answer: \answerNo{}.
    \item[] Justification: Following common practice in stroke-based rendering and image reconstruction benchmarks, the experiments report aggregate quantitative metrics on fixed evaluation sets together with qualitative comparisons, renderer analysis, and ablations across multiple baselines. These analyses are sufficient to support the main conclusions, but the paper does not additionally report error bars, confidence intervals, or statistical significance tests.

\item {\bf Experiments compute resources}
    \item[] Question: For each experiment, does the paper provide sufficient information on the computer resources (type of compute workers, memory, time of execution) needed to reproduce the experiments?
    \item[] Answer: \answerYes{}.
    \item[] Justification: The experimental setup reports training on 4 NVIDIA RTX 4090 GPUs, 200 epochs, and batch size 32, while the supplementary material reports an 8-hour training time and a 0.15-second initialization pipeline for $1024\times1024$ images on a single RTX 4090. Tables also report average per-stroke timing and renderer forward/backward costs.
    
\item {\bf Code of ethics}
    \item[] Question: Does the research conducted in the paper conform, in every respect, with the NeurIPS Code of Ethics \url{https://neurips.cc/public/EthicsGuidelines}?
    \item[] Answer: \answerYes{}.
    \item[] Justification: The research uses public image datasets for stroke-based rendering and does not involve sensitive personal data, surveillance, biometric identification, or safety-critical deployment. The user study is limited to perceptual ratings of rendered images and is described anonymously in the supplementary material.

\item {\bf Broader impacts}
    \item[] Question: Does the paper discuss both potential positive societal impacts and negative societal impacts of the work performed?
    \item[] Answer: \answerYes{}.
    \item[] Justification: As discussed in the supplementary material (Sec.~\ref{sec:broader_impacts}), the approach offers positive societal benefits for photo-to-art creation, beginner painting education, and future controllable generative models. The same section notes that no direct high-risk negative societal impact is identified, while also emphasizing responsible-use boundaries concerning image rights, authorship, and consent.
    
\item {\bf Safeguards}
    \item[] Question: Does the paper describe safeguards that have been put in place for responsible release of data or models that have a high risk for misuse (e.g., pre-trained language models, image generators, or scraped datasets)?
    \item[] Answer: \answerNA{}.
    \item[] Justification: The paper does not release a high-risk generative model, scraped dataset, pretrained language model, or other asset requiring controlled release safeguards. The proposed method is a stroke-based rendering framework evaluated on cited image datasets.

\item {\bf Licenses for existing assets}
    \item[] Question: Are the creators or original owners of assets (e.g., code, data, models), used in the paper, properly credited and are the license and terms of use explicitly mentioned and properly respected?
    \item[] Answer: \answerYes{}.
    \item[] Justification: The paper cites the datasets and baselines used in the experiments. Suppl.~\ref{tab:dataset_licenses} reports the available license or usage terms for Flickr2K, DIV2K, and Im2Oil Gallery, and states that the original images are used only for academic research and are not redistributed.

\item {\bf New assets}
    \item[] Question: Are new assets introduced in the paper well documented and is the documentation provided alongside the assets?
    \item[] Answer: \answerNA{}.
    \item[] Justification: The current submission introduces a method but does not release a new dataset, benchmark, model checkpoint, or other standalone asset requiring separate documentation alongside the asset.

\item {\bf Crowdsourcing and research with human subjects}
    \item[] Question: For crowdsourcing experiments and research with human subjects, does the paper include the full text of instructions given to participants and screenshots, if applicable, as well as details about compensation (if any)? 
    \item[] Answer: \answerYes{}.
    \item[] Justification: The supplementary user-study section reports the number of participants, randomized presentation protocol, rating criteria, participant instruction text, informed-consent procedure, anonymous data collection, and compensation status.

\item {\bf Institutional review board (IRB) approvals or equivalent for research with human subjects}
    \item[] Question: Does the paper describe potential risks incurred by study participants, whether such risks were disclosed to the subjects, and whether Institutional Review Board (IRB) approvals (or an equivalent approval/review based on the requirements of your country or institution) were obtained?
    \item[] Answer: \answerYes{}.
    \item[] Justification: The supplementary user-study section states that the study was an anonymous, minimal-risk perceptual survey approved by the authors' institutional review process for anonymous perceptual surveys. It also describes informed consent, voluntary participation, the absence of personally identifiable or sensitive data collection, and the use of anonymous ratings for academic research.

\item {\bf Declaration of LLM usage}
    \item[] Question: Does the paper describe the usage of LLMs if it is an important, original, or non-standard component of the core methods in this research? Note that if the LLM is used only for writing, editing, or formatting purposes and does \emph{not} impact the core methodology, scientific rigor, or originality of the research, declaration is not required.
    \item[] Answer: \answerNA{}.
    \item[] Justification: The core method consists of differentiable B\'ezier splatting, a feedforward stroke decoder, and stroke-level optimization/auditing. It does not use LLMs as an important, original, or non-standard component of the research methodology.

\end{enumerate}

\end{document}